\documentclass{article}

\usepackage{microtype}
\usepackage{graphicx}
\usepackage{subfigure}
\usepackage{booktabs} % for professional tables
\usepackage{hyperref}

\usepackage[accepted]{icml2024}
\usepackage{amsmath}
\usepackage{amssymb}
\usepackage{mathtools}
\usepackage{amsthm}
\usepackage{pifont}

\usepackage[capitalize,noabbrev]{cleveref}
\usepackage{bbm}
\usepackage{algorithm}
\usepackage{algorithmic}
\renewcommand{\algorithmicrequire}{\textbf{Input:}}
\renewcommand{\algorithmicensure}{\textbf{Output:}}
\usepackage{multirow}
\usepackage{float}
\usepackage{enumitem}
\usepackage{graphicx}
\usepackage[font=small,skip=10pt]{caption}
\usepackage{stfloats}
\usepackage{tikz}
\usepackage{makecell}

\theoremstyle{plain}

\theoremstyle{definition}

\theoremstyle{remark}

\usepackage[textsize=tiny]{todonotes}
\usepackage{soul}
\usepackage{longtable}

\icmltitlerunning{Adaptive Text Watermark for Large Language Models}

\begin{document}

\twocolumn[
\icmltitle{Adaptive Text Watermark for Large Language Models}

\begin{icmlauthorlist}
\icmlauthor{Yepeng Liu}{y}
\icmlauthor{Yuheng Bu}{y}
\end{icmlauthorlist}

\vskip 0.08in

% \begin{icmlauthorlist}
% {\texttt{\{yepeng.liu, buyuheng\}@ufl.edu}}{}
% \end{icmlauthorlist}

\icmlaffiliation{y}{University of Florida}
\icmlcorrespondingauthor{Yepeng Liu}{yepeng.liu@ufl.edu}
\icmlcorrespondingauthor{Yuheng Bu}{buyuheng@ufl.edu}
\icmlkeywords{Machine Learning, ICML}

\vskip 0.3in
]



\printAffiliationsAndNotice{}  % leave blank if no need to mention equal contribution

\begin{abstract}
The advancement of Large Language Models (LLMs) has led to increasing concerns about the misuse of AI-generated text, and watermarking LLM-generated text has emerged as a potential solution.
However, it is challenging to generate high-quality watermarked text while maintaining robustness, security, and the ability to detect watermarks without prior knowledge of the prompt and model. This paper proposes an adaptive text watermarking strategy to address such a challenge. 
To improve the text quality and maintain robustness, we adaptively add watermarking to token distributions with high entropy measured by an auxiliary model and keep the low-entropy token distributions untouched.
For the sake of security and to further minimize the watermark's impact on text quality, instead of using a fixed green/red list generated from a random secret key, which can be vulnerable to decryption and forgery, we adaptively scale up the output logits based on the semantic embedding of previously generated text using a well designed semantic mapping model.
Our experiments involving various LLMs demonstrate that our approach achieves comparable robustness performance to existing watermark methods. Additionally, the text generated by our method has perplexity comparable to that of \emph{un-watermarked} LLMs while maintaining sufficient security.\footnote{ Our code is available at \url{https://github.com/yepengliu/adaptive-text-watermark.git}.}

% \renewcommand{\thefootnote}{}
% \footnotetext{Our code will be available at github.com/yepengliu/adaptive-watermark.}

\end{abstract}

\section{Introduction}
\label{Introduction}
The rapid development of Large Language Models (LLMs) \cite{zhang2022opt, touvron2023llama} has ushered in transformative possibilities but also brings forth potential challenges. One primary challenge lies in the misuse of AI-generated text for malicious purposes, such as spreading disinformation, generating fake news, or engaging in plagiarism. The sheer volume and quality of AI-generated text make it increasingly difficult to discern between human and AI-generated content. As a solution, implementing a text watermark on LLMs proves to be an effective method for detecting AI-generated text.

Specifically, the watermarking technique encodes certain patterns into the text generated by LLMs, which are imperceptible to humans but can be detected by the algorithm. This is usually achieved by introducing well-designed perturbations to the original logits or using a special sampling method. \citet{kirchenbauer2023watermark} design a watermark for LLM by randomly separating the vocabulary into `green' and `red' lists and increasing the frequency of tokens from the `green' list. The previously generated token is used as a secret key to determine the `green/red' list for the next token. This method achieves good detection performance when the watermarked text remains unmodified. However, after modifications such as word replacement and paraphrasing, this method will fail as the `green/red' list cannot be reliably recovered from the modified text. As it is common for individuals to alter AI-generated text rather than use it directly, improving the \emph{robustness} of the watermark becomes an essential requirement.

To improve the robustness of LLM watermarking, \citet{zhao2023provable} propose to use a fixed `green/red' list for each token, which is generated from a random secret key shared with the detector. Both empirical and theoretical evidence demonstrate that this method substantially enhances robustness.
However, more recently, \citet{sadasivan2023aigenerated} introduced a spoofing attack specifically designed to deduce hidden watermark patterns by analyzing the token frequency, thereby enabling the forgery of watermarked text. Such an attack offers a method for adversaries to circumvent the watermark, potentially damaging the reputation of the watermarking system.
Regrettably, it has been shown that the algorithm proposed in \citet{zhao2023provable} is vulnerable to spoofing attack, and one can infer the fixed `green' token list and forge the watermarked text.

The aforementioned papers enhance the frequency of tokens in the `green' list during generation by uniformly adding a fixed value to the logits of those `green' tokens. This way of perturbing the distribution may change the distribution significantly, deteriorating the quality of the watermarked text. To tackle this issue, \citet{kuditipudi2023robust} presents a distortion-free watermarking method that maintains the original probability distribution of LLMs and embeds the watermark during the token sampling process. 
% This method is able to generate watermarked text with perplexity close to that of \emph{un-watermarked} text. 
However, such a distortion-free watermarking lacks sufficient robustness against modifications, like paraphrasing.

Inspired by existing research, designing an effective watermark algorithm requires us to consider at least the following three factors comprehensively. (1) \textbf{Robustness:} The watermark should remain detectable after modifications without significantly changing the semantics of the watermarked text. (2) \textbf{Security:} The watermark should be resilient against efforts to reverse-engineer or decryption without authorization. (3) \textbf{Text quality:} LLMs-generated watermarked text should preserve a level of perplexity comparable to the un-watermarked text. Among existing works, achieving high-quality text often comes with the trade-off of robustness, and prioritizing strong robustness can potentially compromise security. Therefore, designing a watermarking scheme that simultaneously guarantees robustness, security, and text quality poses a significant challenge.

\begin{figure*}[!t]
    \centering
    \includegraphics[width=1\linewidth]{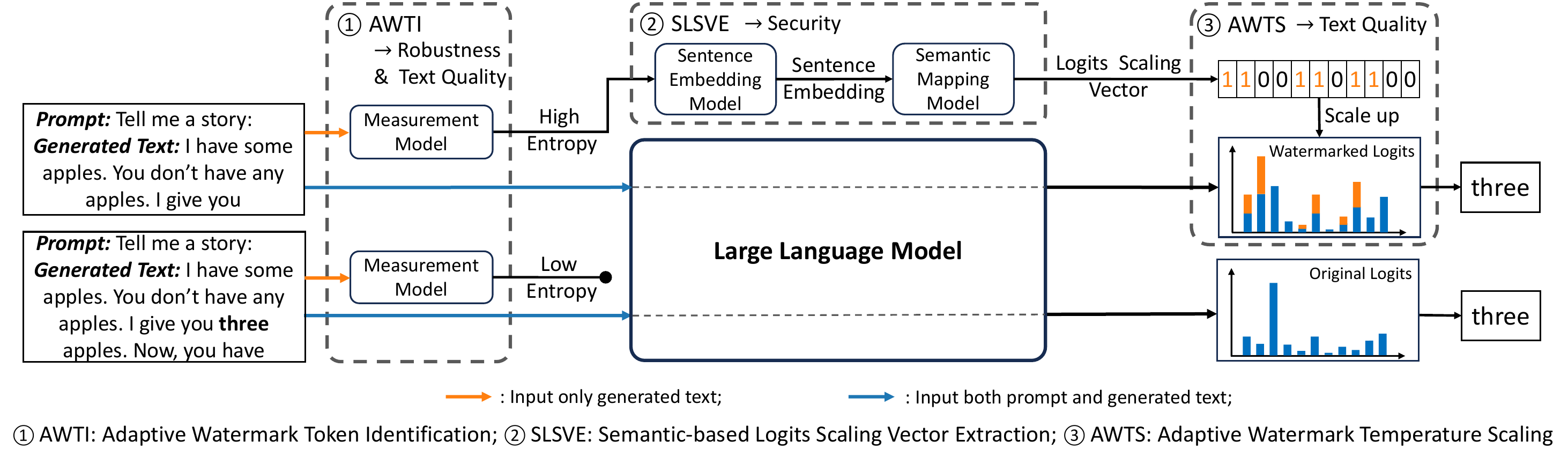}
    \vspace{-1.5em}
    \caption{Workflow of the proposed Adaptive Watermark. Our Adaptive Watermark method will assess the entropy of the distribution for all previously generated tokens and add watermarks only to high-entropy tokens.  This figure illustrates two cases in the text generation process. For the first example, the distribution of the next token has high entropy as measured by AWTI, indicating high uncertainty. Then, the SLSVE module will extract the logits scaling vector based on the semantics of this text. Subsequently, we apply AWTS to perturb the sampling distribution. As for the second text, it results in low entropy measured by AWTI, suggesting a low uncertainty for the next token. Then, the next token will be directly sampled from the original probability distribution. 
    }
    \vspace{-0.5em}
    \label{fig:overvew}
\end{figure*}

In this work, our goal is to provide a holistic approach that integrates robustness and security with the production of high-quality text. Additionally, we ensure the watermark can be detected in an agnostic manner, meaning that the watermark detection process is independent of the original LLMs and prompts. Our watermark strategy consists of three components, as shown in Figure \ref{fig:overvew}. 
\begin{itemize}
[leftmargin=*,topsep=0em,itemsep=-0.3em]
    \item To minimize the impact of the watermark on the text quality while maintaining the robustness, we propose the Adaptive Watermark Token Identification (AWTI) so that the watermarking is added adaptively to high-entropy token distributions and leaves the low-entropy token distributions intact.
    This method employs an auxiliary Measurement Model to assess the entropy of each token,  ensuring that the entropy can be reliably estimated without access to the original LLMs and prompts.

    \item To further enhance the security of our method, we introduce the Semantic-based Logits Scaling Vector Extraction (SLSVE).
    % to replace the widely used `green/red' list. 
    Our logits scaling vector is obtained by passing the semantics of the generated text (using a pre-trained Sentence Embedding Model) to a Semantic Mapping Model, which is hard to reverse-engineer.

    \item We propose the Adaptive Watermark Temperature Scaling (AWTS), which proportionally scales up the original logits using the extracted logits scaling vector. Through this approach, the perturbation is applied by scaling the \emph{temperature} of the distribution, which further reduces the influence on the text quality due to watermarking.
    
    \item  In the detection process, the AWTI is first applied to the text under analysis to identify potential \emph{watermarked tokens}. After this, the SLSVE is employed to extract the logits scaling vectors and calculate the detection score. 
\end{itemize}

We conduct extensive experiments over various language models (OPT-6.7B~\cite{zhang2022opt}, GPT-J-6B~\cite{gpt-j}, and Mistral-7B \cite{jiang2023mistral}) to show the effectiveness of our method. We employ perplexity as a metric to evaluate text quality, and our results indicate that the quality of the watermarked text remains comparable to that of the \emph{un-watermarked} text. To show the robustness of our method, we paraphrase the watermarked text using GPT3.5 and DIPPER \cite{krishna2023paraphrasing}. The outcomes are on par with those achieved by \citet{zhao2023provable}, which is the most robust method in the literature. The spoofing attack is used to illustrate that our watermark is hard to decrypt. Lastly, our detection process is agnostic and does not rely on any original prompt and LLMs.

\section{Related Work}
As LLMs evolve, the need to distinguish AI-generated text from human-written text is crucial for authenticity, accountability, and intellectual property protection. Text watermarking, embedding hidden markers in text, emerges as a promising solution for this challenge.
There are currently two main types of watermark methods in use: generated text watermarking and LLMs watermarking.

\textbf{Generated text watermarking} primarily operates on existing text, altering aspects like format \cite{brassilformat, por2012unispach,sato2023embarrassingly, stefano2016unicode}, lexical choices \cite{yoo2023robust, yang2023watermarking, munyer2023deeptextmark, yang2021tracing}, or syntax \cite{atallah2001ih, topkara2006slnlw, MERAL2009107}, while preserving the underlying semantics. EASYMARK \cite{sato2023embarrassingly}, a format-based watermarking, utilizes subtle changes in Unicode characters and spacing to embed watermarks in text. For word-level modifications, \citet{yang2021tracing}, based on context-aware lexical substitution, utilizes BERT to identify semantically suitable synonym candidates for watermarking. \citet{MERAL2009107} introduces a morphosyntax-based natural language watermarking scheme that uses syntactic tree diagrams to analyze the structure and functional dependencies of text. In addition to the aforementioned methods that primarily rely on specific rules and might lead to unnatural modifications, some neural network-based approaches \cite{zhang2023remarkllm, abdelnabi2021adversarial} add watermarking by regenerating the text with specific message signatures using a pre-trained language model, while keeping the semantics unchanged. 

\textbf{LLMs watermarking.} Rather than adding a watermark to existing text, an increasing number of studies are focusing on embedding watermarks directly into the text during the LLM generation \cite{wang2023towards, fairoze2023publicly,hu2023unbiased, lee2023wrote, huo2024token, fu2024gumbelsoft, zhao2024permute, zhang2023watermarks, tu2023waterbench}. 

Some research efforts are dedicated to enhancing the \emph{robustness} of watermarks \cite{kirchenbauer2023watermark, kirchenbauer2023reliability, liu2023semantic, zhao2023provable, ren2023robust}, aiming to ensure the persistence of watermarking even after exposure to various attacks. \citet{zhao2023provable} introduce a provable robust watermark using a fixed vocabulary `green/red' list generated using a shared random key. \citet{ren2023robust} utilize the semantics of the generated text as a hash key to separate the vocabulary, which is achieved by discretizing the continuous embedding space. \citet{hou2023semstamp} propose a sentence-level semantic watermarking technique that utilizes the semantics of previously generated sentences to divide the semantic space into ``valid" and ``blocked" regions and generates new sentences through rejection sampling until its semantics reside within the ``valid" region. 

Some works focus on enhancing the \emph{security} of watermarks \cite{wu2023dipmark, christ2023undetectable, liu2023unforgeable}, aiming to make them difficult to forge. \citet{liu2023semantic} proposed a semantic invariant watermark that balances robustness and security. However, this method relies on the original prompts for detection, which are often impractical to access. Additionally, the way it perturbs the distribution results in noticeably higher text perplexity compared to un-watermarked text. 

All the watermarks mentioned above are inserted into the text by manipulating the logits produced by LLMs, which inevitably introduces some distortion in text generation. Some studies \citet{kuditipudi2023robust, christ2023undetectable}  introduce watermarks using specially crafted sampling methods. These methods are designed to ensure that the watermark has a minimal distortion on the original LLM distributions to improve \emph{text quality}.

Our work focuses on the LLMs watermarking. Previous works in this field have significantly improved the robustness, text quality, and security aspects of watermarks in a separate manner. However, it remains challenging to consider all these aspects simultaneously. Therefore, we intend to propose a holistic watermark approach that achieves strong robustness and high-quality text generation while also being difficult to forge. 

% \section{Preliminary}
\textbf{Notations.} Given a Language Model (LM), denote $\mathcal{V}$ as the vocabulary of LM, and denote $\mathcal{|\mathcal{V}|}$ as the size of vocabulary. For a input prompt $R=\{R^{(1)}, R^{(2)}, ..., R^{(N)}\}$, the LM will generate a sequence of tokens $S=\{S^{(1)}, S^{(2)}, ..., S^{(T)}\}$ based on the prompt. We denote $S_{0:t}=\{S^{(1)}, ..., S^{(t)}\}$ and $S_0=\emptyset$, and $r^{(n)}$ and $s^{(t)} $ are concrete tokens in the vocabulary, i.e., $r^{(n)}$, $s^{(t)}\in\mathcal{V}$. As an auto-regressive model, LM will predict the probability distribution $P_{LM}(\cdot|R,S_{0:t})$ over each token in $\mathcal{V}$ at time step $t+1$ based on the prompt and the preceding sequence, and sample the next token based on this probability distribution until the token of termination is sampled.

\section{Adaptive Watermark Generation}
In this section, we present our adaptive watermarking technique designed to minimize the impact on text quality, enhance security, and maintain robustness simultaneously. Section \ref{AWTI} introduces Adaptive Watermark Token Identification to ensure watermark robustness while minimizing the impact of text quality in generated content. The Semantic-based Logits Scaling Vector Extraction, as discussed in Section \ref{SLSVE}, generates different logits scaling vectors based on the semantics of the generated text, which adaptively perturb the distribution of the LLM.
% replaces the widely used ‘green/red’ list to separate the vocabulary, making it more difficult to forge. 
In Section \ref{AWTS}, we propose a new distribution perturbation method to inject watermarking, further improving watermarked text quality.

\subsection{Adaptive Watermark Token Identification (AWTI)}
\label{AWTI}
The Adaptive Watermark Token Identification (see Figure~\ref{fig:overvew}-\ding{192}) is designed to minimize the impact of the watermark on the text quality while maintaining its robustness by reducing the percentage of perturbed tokens during the text generation process. The watermarking is added in an adaptive manner, i.e., only the distribution of those tokens flagged by our auxiliary language model as having high entropy will be perturbed by our watermarking method.

Specifically, given a prompt $R$, the LLM will generate a logits vector $l^{(t+1)}$ over all the tokens in $\mathcal{V}$ at time step $t+1$. The \textit{softmax} function is then applied to convert the logits into a probability distribution $P_{LM}(\cdot|R,S_{0:t})$, where the probability of the $k$-th token in the $\mathcal{V}$ is $p^{(t+1)}_k={\rm exp}(l^{(t+1)}_k)/\sum_{i=1}^{|\mathcal{V}|} {\rm exp}(l^{(t+1)}_i).$ The LLM will sample a token from this probability distribution using methods such as Top-$K$ sampling or Top-$p$ sampling. As shown in the example of Figure~\ref{fig:overvew}, the sampling distribution may vary based on the context of the generated text at different steps. 

Shannon Entropy ~\cite{Shannon} measures the uncertainty of a distribution, which is defined as $H(P_X)\triangleq-\sum_{x \in \mathcal{X}} P_X(x){\log} P_X(x)$,
% \begin{equation}
%     H\left(X\right)=-\sum_{i=1}^{n}P(x_i){\log}P(x_i),
%     \nonumber
% \end{equation}
where $P_X$ is a discrete probability distribution defined over the space $\mathcal{X}$, and $0\leq H(P_X)\leq \log|\mathcal{X}|$.  When the entropy is low, the distribution becomes more concentrated on specific tokens. Conversely, when the entropy is high, the distribution becomes more spread, indicating larger uncertainty. A high entropy distribution in LLMs means that the original distribution offers multiple reasonable choices during text generation. Thus, inserting certain patterns by only perturbing the high entropy distribution will minimize the impact of the sampling process, ensuring that the generated tokens remain contextually appropriate.

We note that \citet{lee2023wrote} also discusses the strategy of perturbing only the token with high entropy distributions for code generation. Their approach, however, requires access to both the original LLM and prompts, rendering it impractical for watermarking detection. Additionally, utilizing the original LLM for detection is not only time-consuming but also incurs substantial computational costs. 

To ensure reliable entropy estimation during the text generation and detection process without relying on the watermarked LLM or the original prompts, we use an auxiliary Measurement Model (MM) to compute the entropy of the generated text at each time step. 
The entropy measured with MM is expressed as
\begin{equation}
    H\left(P_{MM}(\cdot|S_{0:t})\right)=-\sum_{i=1}^{|\mathcal{V}|}p^{(t+1)}_i \log p^{(t+1)}_i.
\end{equation}
The MM can be a language model that is significantly smaller in scale compared to the watermarked LLMs, such as GPT-2~\cite{radford2019gpt2} or OPT-350M~\cite{zhang2022opt}. 
We use a predefined entropy threshold, denoted as $\alpha$, 
% to avoid disturbing distributions whose entropy is below this threshold. 
% Therefore, 
and our watermark only perturbs those distributions with $H\left(P_{MM}(\cdot|S_{0:t})\right)\geq \alpha$. We refer to the token sampled from the intact distribution as \emph{un-watermarked token} and the token sampled from the perturbed distribution as \emph{watermarked token}.
As the threshold $\alpha$ increases, two effects become evident. Firstly, there will be more un-watermarked tokens during the generation process. Secondly, the impact of each
watermarked token diminishes, mainly due to the higher inherent uncertainty in their distributions. As a result, both effects will enhance the quality of the watermarked text. 

It is important to acknowledge that the entropy computed by the auxiliary MM may differ from that of the original watermarked LLM, but this discrepancy does not affect the efficacy of our watermarking. The auxiliary MM is employed primarily to assist in identifying potential watermarked tokens during the detection phase, eliminating the need for the original watermarked LLM and prompt. As long as we can identify watermarked tokens in the detection phase using such entropy estimates, the performance will not be affected. In addition, due to the fact that only watermarked tokens are involved in the detection process as described in Section~\ref{adaptive_detection}, adaptive watermarking may also benefit robustness by filtering out un-watermarked tokens.

\subsection{Semantic-based Logits Scaling Vector Extraction}
\label{SLSVE}

The Semantic-based Logits Scaling Vector Extraction (see Figure~\ref{fig:overvew}-\ding{193}) is primarily developed to ensure the security of the watermark. The intuition of this approach is to make sure that the perturbation in the distribution is determined by the semantics of the generated text using a well-designed neural network so that the inserted watermarking cannot be easily decrypted by attackers.

Previous work uses the hash of the preceding token or a fixed random key to generate the `green' and `red' lists. However, such lists can be recovered by analyzing the token frequency in watermarked text using statistical approaches, e.g., spoofing attacks, which enable adversaries to forge watermarks.  In contrast, the proposed SLSVE will adaptively generate the logits scaling vector to perturb the distribution based on the semantics of the generated text using a trainable neural network. 
The complex high-dimensional semantic space and neural networks further increase the complexity of the watermark, making it more difficult for adversaries to decrypt.

\begin{algorithm}[t!]
    \caption{Semantic-based Logits Scaling Vector Extraction (SLSVE)}
    \begin{algorithmic}[1]
        \REQUIRE{Sentence Embedding Model ($\rm SE$), Semantic Mapping Model ($\rm SM$), Input sentence $S$.}
        \vspace{3pt}
        \STATE $u^{(t)}(S)={\rm SE}(S_{0:t-1})$
        \STATE $v^{(t)}(S)={\rm SM}(u^{(t)}(S);\theta)$
        \STATE $\hat{v}^{(t)}(S)=\big[\mathbbm{1}\{v_1^{(t)}(S)>0\},...,\mathbbm{1}\{v_{|\mathcal{V}|}^{(t)}(S)>0\}\big]$
        \ENSURE logits scaling vector $\hat{v}^{(t)}$.
    \end{algorithmic}
\end{algorithm}
In particular, for tokens with distributions satisfying $H(P_{MM}(\cdot|S_{0:t-1}))\geq\alpha$ at step $t$, we proceed to extract logits scaling vector based on the semantics of generated text. 
First, we obtain the semantic embedding using a Sentence Embedding ($\rm SE$) Model, i.e., $u^{(t)}(S)={\rm SE}({S}_{0:t-1})$, where $u^{(t)}\in \mathbbm{R}^L$. We employ a pre-trained LM as the Sentence Embedding Model, e.g., Sentence-BERT \cite{reimers2019sentencebert}, to extract the sentence embedding. Then, we train a Semantic Mapping ($\rm SM$) Model to convert the sentence embedding into a logits scaling vector, denoted as $v^{(t)}(S)={\rm SM}(u^{(t)}(S); \theta)$, where $v^{(t)}\in\mathbbm{R}^{|\mathcal{V}|}$ and $\theta$ denotes the parameters of the Semantic Mapping Model.

Specifically, we train a feedforward neural network with multiple residual connections to ensure that the logits scaling vector satisfies the following properties.

(1) \textbf{Smoothness with respect to semantics:} As we need to recover the logits scaling vector to perform detection even with modifications in the watermarked text, the Semantic Mapping Model needs to be insensitive to small changes in semantics to improve the robustness of the watermark. 
In order to make the logits scaling vector align with the change in semantic embedding, we want to make sure that if two different sentences $S$ and $S'$ are close in embedding space, they will have similar logits scaling vector $v$ and $v'$. We will minimize the loss $\big|D(u(S), u(S'))-D(v(S),v(S'))\big|$, where we use the Euclidean distance $D(a,b)=\sqrt{\sum_{i=1}(a_i-b_i)^2}$ to measure the distance between two sentence embeddings.

(2) \textbf{Uniform Perturbation:} To make the watermark easy to detect, we want to ensure the logits scaling vector increases the probability of roughly half of the tokens in the vocabulary. For any sentence $S$ within the training dataset $\mathcal{T}$, we ensure that $\sum_{i \in \mathcal{V}}\mathrm{sign}\big(v_i(S)\big)=0$. Therefore, an equal rate of positive and negative entries in $v_i(S)$.

(3) \textbf{Unbiased token preference:} The logits scaling vector should be designed to prevent consistent bias towards specific tokens in the vocabulary. This ensures that specific tokens do not always exhibit a high probability in distributions across various semantic contexts, preventing their frequent recurrence in the watermarked text and improve security.
% so that cannot be detected by the spoofing attack.  
Therefore, for any token $i$ in the vocabulary $\mathcal{V}$, we will ensure $\sum_{S\in\mathcal{T}}\mathrm{sign}\big(v_i (S)\big)=0$.

Additionally, during the training of the Semantic Mapping Model, we ensure that the logits scaling vector remains stable against semantic variations using the following three implementation tricks. This further guarantees that the watermark's robustness remains unaffected by semantic variations within a specific range. (i) Due to the high-dimensional nature of sentence embeddings, experiments on the original training dataset revealed that most sentence embeddings are essentially orthogonal to each other.
To improve the model's capability in effectively handling slight semantic perturbations,  we enlarged our training dataset by performing data augmentation, including paraphrasing, expanding, and shortening the sentences in the original training dataset. (ii) With the augmented data mentioned above, a contrastive loss term is applied during training to minimize the distance between the sentence ($S\in\mathcal{T}$) and its augmented counterparts($\widetilde{S}\in\widetilde{\mathcal{T}}$), i.e., $\min \sum_{S\in\mathcal{T}, \widetilde{S}\in\widetilde{\mathcal{T}}} \big| D(v(S), v(\widetilde{S})) \big|$. This contrastive term improves the stability of the model when a modified sentence is presented. (iii) We rescale the Euclidean distance between two sentence embeddings $d=D(u(S), u(S')) \in[L, U]$ to a wider range $d'\in[L',U']$ using a linear transformation, where $L'<L$ and $U'>U$.
Rescaling makes the distance between two sentence embeddings smaller when they are close to each other and increases the distance when they are originally farther apart in the embedding space.

Therefore, considering all the desired properties, the loss function of the Semantic Mapping Model is:
\begin{align} \label{loss_function}
    \mathcal{L}(\widetilde{\mathcal{T}}, \theta)=&\sum_{S, S'\in \widetilde{\mathcal{T}}}\!\big|D(u(S), u(S'))\!-\!D(v(S),v(S'))\big| \nonumber \\
    +&  \!\sum_{S\in\widetilde{\mathcal{T}}}\!\big|\!\sum_{i\in|\mathcal{V}|} \mathrm{sign}(v_i(S))\big| 
    \!+\!\sum_{i\in|\mathcal{V}|}\!\big|\!\sum_{S\in\widetilde{\mathcal{T}}} \mathrm{sign}(v_i (S))\big| \nonumber\\
    +& \sum_{S\in\mathcal{T}, \widetilde{S}\in\widetilde{\mathcal{T}}} \big| D(v(S), v(\widetilde{S})) \big|. 
\end{align}
After getting the logits scaling vector $(v^{(t)})$ using the trained model, we apply the indicator function to transfer every entry in $v^{(t)}$ to either $0$ or $1$, i.e., $\hat{v}_i^{(t)}(S) = \mathbbm{1}\{v_i^{(t)}(S)>0\}$.

\subsection{Adaptive Watermark Temperature Scaling (AWTS)}
\label{AWTS}
To further minimize the impact of perturbation on the original probability distribution, we introduce the Adaptive Watermark Temperature Scaling method (see Figure~\ref{fig:overvew}-\ding{194}). This technique enhances the consistency of the perturbed distribution with the original distribution by adaptively adjusting the \emph{temperature} of the distribution.

Specifically, we add watermark to the original LLM logits $l^{(t)}$ using temperature scaling:
\begin{equation}
    \hat{l}^{(t)}=l^{(t)} \cdot (1+\delta \cdot \hat{v}^{(t)}),
\end{equation}
where $\delta>0$ controls the strength of the watermark. In this way, we adaptively perturb the logit of each token by proportionally scaling up a factor of $(1+\delta \cdot \hat{v}^{(t)})$ using the obtained SLSVE. 
If the original logit $l_i^{(t)}$ is small, the $\hat{l}_i^{(t)}$ after the manipulation remains small, indicating that our watermark slightly perturbs these logits. Temperature scaling mainly affects tokens with large original logits $l_i^{(t)}$, causing the corresponding $\hat{l}_i^{(t)}$ to increase more significantly. In contrast to uniformly increasing the logits of tokens in the `green list' as in previous work \cite{kirchenbauer2023watermark}, our approach inserts a watermark by modulating the \textit{temperature} of the logits. Therefore, our watermark strength adaptively depends on the original distribution, which will make the generated watermarked text more coherent. As we will show in Figure \ref{fig:pplx_ab}, AWTS also contributes to the improvement of text quality.

\begin{algorithm}[t!]
    \caption{Adaptive Watermark Generation}
    \begin{algorithmic}[1]\label{alg2}
        \REQUIRE{Language model $P_{LM}$, measurement model $P_{MM}$, prompt $R$, watermark strength $\delta$, opening sentence $\hat{S}$, entropy threshold $\alpha$, measure threshold $M$.}
        \vspace{3pt}
        \FOR {$t=1,2,...,T$}
        \IF {$t\leq M$}
        \STATE $\hat{v}^{(t)} = \mathrm{SLSVE}(\hat{S}) $
        \STATE $\displaystyle q_k^{(t)}=\frac{\exp(l_k^{(t)}\cdot (1+\delta \cdot \hat{v}_k^{(t)}))}{\sum_{i\in \mathcal{V}} \exp(l_i^{(t)} \cdot (1+\delta \cdot \hat{v}_i^{(t)}))} $
        \vspace{3pt}
        \STATE $\displaystyle s^{(t)}\sim q^{(t)}$.
        \ELSE
        \IF{$H(P_{MM}(\cdot|S_{0:t-1})\geq \alpha$}
        \STATE $\hat{v}^{(t)} = \mathrm{SLSVE}(S_{0:t-1}) $
        \STATE $\displaystyle q_k^{(t)}=\frac{\exp(l_k^{(t)}\cdot (1+\delta \cdot \hat{v}_k^{(t)}))}{\sum_{i\in \mathcal{V}} \exp(l_i^{(t)} \cdot (1+\delta \cdot \hat{v}_i^{(t)}))} $
        \vspace{3pt}
        \STATE $\displaystyle s^{(t)}\sim q^{(t)}$.
        \ELSE
        \STATE
        $\displaystyle p_k^{(t)}=\frac{\exp(l_k^{(t)})}{\sum_{i\in \mathcal{V}} \exp(l_i^{(t)})}$,
        \vspace{3pt}
        \STATE
        $\displaystyle s^{(t)}\sim p^{(t)}$.
        \ENDIF
        \ENDIF
        \ENDFOR
        \vspace{3pt}
        \ENSURE watermarked sequence $s^{(1)}, s^{(2)}, ..., s^{(T)}$.
    \end{algorithmic}
\end{algorithm}

Algorithm~\ref{alg2} describes the entire procedure of the proposed Adaptive Watermark algorithm.
At the beginning of the text generation process, only a limited number of tokens can be used for extracting semantic embeddings, which makes the detection of the watermark unstable under attack. To maintain the stability and robustness of the watermark, we set a secret opening sentence $\hat{S}$. When the number of generated tokens is smaller than a pre-determined measure threshold $M$, we extract the semantic embedding from the opening sentence and apply SLSVE.

\section{Adaptive Watermark Detection}
\label{adaptive_detection}
In this section, we propose to approximate the Likelihood Ratio Test (LRT) to detect the watermark. Consider the following null hypothesis $H_0$ and the alternative $H_1$,
\begin{align*}
    &H_0: {\rm The\ candidate\ text\ is\ not\ watermarked.}  \\
    &H_1: {\rm The\ candidate\ text\ is\ watermarked.} 
\end{align*}
Given a sequence of tokens $S=\{S^{(1)}, S^{(2)}, ..., S^{(T)}\}$, we input the sequence to MM to calculate the entropy and select the potential \emph{watermarked token} at each time step,  following the same process as the AWTI described in Section \ref{AWTI}. Denote the set of potential watermarked token as $\mathcal{W}$, i.e., for all $s^{(t)} \in \mathcal{W}$, we have $H(P_{MM}(\cdot|S_{0:t-1}))\geq\alpha$. For any $s^{(t)} \in \mathcal{W}$, if it is sampled from the perturbed distribution, the distribution would be $P_{H_1}(\cdot|R, S_{0:t})$. Conversely, if it is sampled from the original distribution, the distribution would be $P_{H_0}(\cdot|R, S_{0:t})$. Subsequently, the logits scaling vector is extracted utilizing both the Semantic Embedding Model and the Semantic Mapping Model, as detailed in Section \ref{SLSVE}. As we perturb the distribution with temperature scaling, the likelihood ratio  can be written as
\begin{align}
    &{\log} ~\frac{P_{H_1}(S^{(t)}=k|R, S_{0:t-1})}{P_{H_0}(S^{(t)}=k|R, S_{0:t-1})} \nonumber \\
    &=~{\log} ~\frac{e^{l_k\cdot(1+\delta \cdot \hat{v}_k)}}{\sum_i e^{l_i\cdot(1+\delta \cdot \hat{v}_i)}}+{\log} ~\frac{e^{l_k}}{\sum_i e^{l_i}}  \nonumber \\
    &=~ l_k\cdot(\delta \cdot \hat{v}_k) + {\log}~ \frac{\sum_i e^{l_i}}{\sum_i e^{l_i\cdot(1+\delta \cdot \hat{v}_i)}}\nonumber \\
    &\approx~ l_k\cdot(\delta \cdot \hat{v}_k),
\end{align}
where we omitted the superscript of time $t$, and in the last step, we approximated the ratio of the normalization factor in the softmax function by 1.

\begin{figure}
    \centering
    \includegraphics[width=1\linewidth]{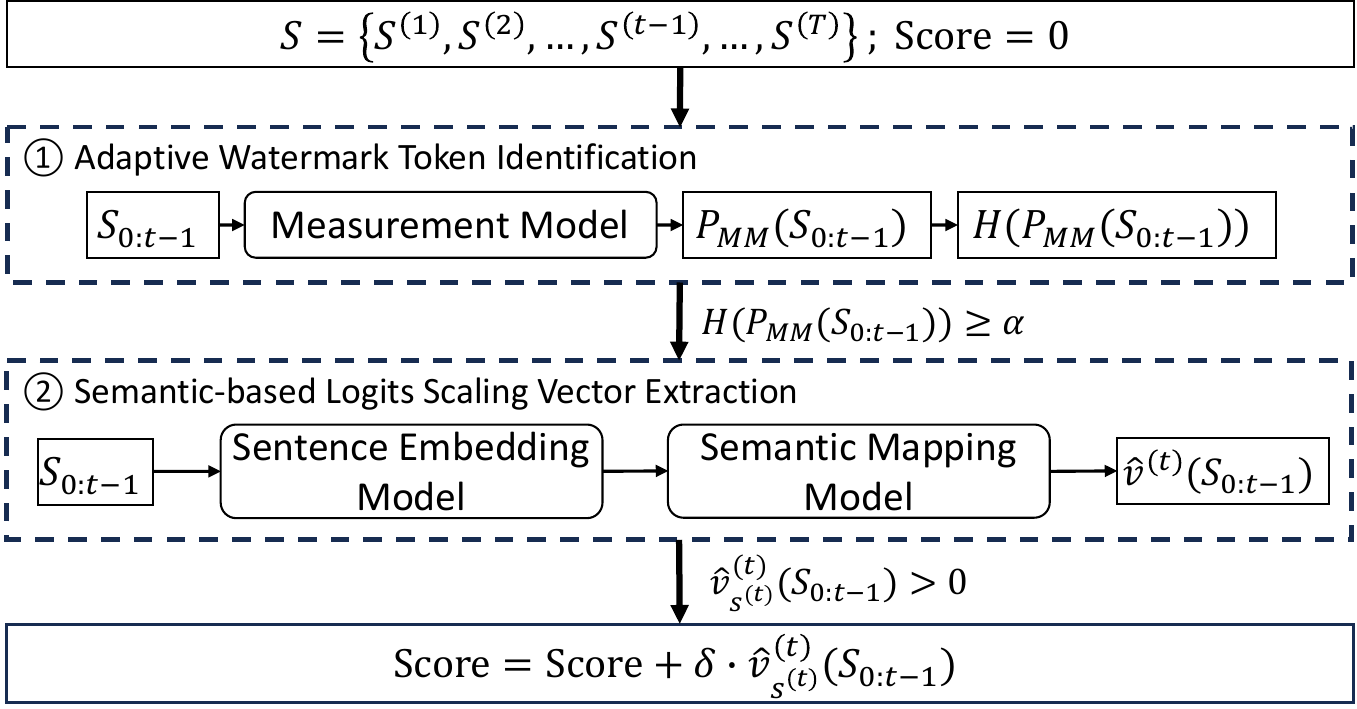}
    % \vspace{-0.5cm}
    \caption{Workflow of Adaptive Detection. This figure illustrates a single step during the detection process. At each time step, we will first check if the current token is a potential \emph{watermarked token} by applying AWTI to estimate the entropy of the current distribution. For potential \emph{watermarked token}, the preceding text will be used to extract the logits scaling vector through SLSVE. If the corresponding value of the potential \emph{watermarked token} in the logits scaling vector is positive, it will be added to the total score.}
    \label{fig:detection_workflow}
    % \vspace{-0.5em}
\end{figure}

In the detection process, the original logits $l_k$ are unknown, and our focus is only on identifying which tokens in the logits have been scaled up.
For this purpose, we simply use $\delta \cdot \hat{v}_k$ as the test score to facilitate detection. 
% our method requires only the extraction of the logits scaling vector to facilitate detection. 
Therefore, the test score is given by
\vspace{-1em}
\begin{align}
    {\rm Score}(S = \{s^{(1)},...,s^{(T)}\})=\frac{\sum_{s^{(t)}\in \mathcal{W}} \delta \cdot \hat{v}^{(t)}_{s^{(t)}}}{|\mathcal{W}|},
\end{align}
where $|\mathcal{W}|$ represents the number of all potential watermarked tokens.

\section{Experiments}
\subsection{Experiment Setting}

\textbf{Implementation Details.} 
In our watermarking algorithm, the default hyperparameters are set as follows: $\alpha=2$, $M=50$. We use Sentence-Transformers \cite{reimers2019sentencebert} to extract the sentence embedding. The specific model we used is \texttt{all-mpnet-base-v2} \footnote{https://huggingface.co/sentence-transformers/all-mpnet-base-v2}. More details for the Semantic Mapping Model are provided in Appendix~\ref{app:exp}. We apply our method to three different models: OPT-2.7B, OPT-6.7B \cite{zhang2022opt}, GPT-J-6B \cite{gpt-j}, and Mistral-7B \cite{jiang2023mistral} (in Appendix~\ref{app:more_model}). At each time step, the next token is sampled from the distribution using a combination of Top-$K$ and Top-$p$ sampling methods, with $K = 50$ and $p=0.9$. 
It is worth noting that the watermarking strength $\delta$ should be chosen differently for different LLMs based on their specific capabilities. For example, we set $\delta=1.5$ for OPT-6.7B and set $\delta=0.6$ for Mistral-7B.

\textbf{Baselines.} Our method is compared with existing methods, including KGW-$0$ \cite{zhao2023provable}, KGW-$1$/KGW-$2$ \cite{kirchenbauer2023watermark}, EXP-edit \cite{kuditipudi2023robust}. In KWG-$1$/KWG-$2$, the numeral denotes the hashing of the previous 1 or 2 tokens, respectively. For KGW-$0$, KGW-$1$, and KGW-$2$, the `green' list token percentage is set to be $0.5$. Additional comparisons against two model non-agnostic baselines, i.e., SWEET~\cite{lee2023wrote} and SIR~\cite{liu2023semantic}, can be found in Appendix~\ref{app:sweet_sir}.

\textbf{Dataset and Prompt.} To test the performance of our methods, the results on \texttt{realnewslike} subset in C4 dataset \cite{c4Raffel} are provided here, and additional results on ELI5 \cite{fan2019eli5} can be found in Appendix~\ref{app:more_data}. For C4, we use the first two sentences in each text as a prompt for LLMs and the subsequent $200$ tokens as human-generated text. For ELI5, we use the question as a prompt and the corresponding answer as human-generated text. For both datasets, we generate $200 \pm 30$ tokens using the LLMs and prompts. The Semantic Mapping Model is trained using the Sentence-Compression dataset \footnote{https://huggingface.co/datasets/embedding-data/sentence-compression}.

\textbf{Evaluation Methods.} Watermark detection performance is evaluated through the ROC-AUC value, the Best F1 Score, and the ROC curve, which illustrates the true positive rate (TPR) and false positive rate (FPR) at various thresholds. The TPR and FPR indicate the accurate identification of watermarked text and the incorrect recognition of human-written text as being watermarked. The perplexity of the text is calculated using LLama-13B \cite{touvron2023llama}, measuring the quality of the text. In Appendices~\ref{app:more_model} and \ref{app:more_data}, we also use a more powerful LLM, GPT-3, to calculate the perplexity.

\subsection{Robustness}
To demonstrate the robustness of our watermark strategy, we perform a paraphrase attack on the watermarked text using two different paraphrasing methods: GPT-$3.5$ and DIPPER \cite{krishna2023paraphrasing}. For GPT-$3.5$, we use the \texttt{gpt-3.5-turbo-instruct} version, and we employ the same prompt as used in \citet{kirchenbauer2023reliability} to paraphrase the watermarked text. DIPPER is a paraphrase generation model to rewrite the text and evade the watermark detection, and the lexical diversity there is set to be $60$.

The results are summarized in Table \ref{robustness}, which shows that all methods demonstrate effective performance when the watermarked texts remain unaltered. However, when the watermarked text is paraphrased using GPT-$3.5$ and DIPPER, the proposed adaptive watermarking method remains high ROC-AUC and Best F1 Score, surpassing those of the baselines across different language models. The corresponding ROC curves, showcasing the performance of different watermarking methods on various language models under paraphrasing attacks, can be found in Appendix~\ref{appendix_additional_roc}. Moreover, we evaluate the TPR scores at fixed 1\% and 10\% FPR as detailed in Appendix~\ref{sub:lowfpr}.

\begin{table*}[]\scriptsize
\centering
\setlength{\tabcolsep}{5pt}
\caption{Performance evaluation of watermarked text under different conditions, including cases with no attack and paraphrase attacks.}\label{robustness}
% \vspace{-0.2cm}
\begin{tabular}{clcccccccccc}
\toprule
\multirow{2}{*}{Language Model} 
& \multirow{2}{*}{Setting} 
& \multicolumn{5}{c}{ROC-AUC}     
& \multicolumn{5}{c}{Best F1Score} 
\\ \cmidrule(lr){3-7}  \cmidrule(lr){8-12}
                                &                         & KGW-0 & KGW-1 & KGW-2 & EXP-edit & \textbf{Ours} & KGW-0  & KGW-1 & KGW-2 & EXP-edit & \textbf{Ours}  \\ \midrule
\multirow{3}{*}{OPT-$2.7$}        & No Attack                &$0.996$       &$0.998$   &$0.998$    &$0.997$          &$\textbf{1.000}$      &$0.989$        &$0.997$   &$0.999$    &$0.994$          &$\textbf{1.000}$      \\
                                & GPT-3.5                  &$0.971$       &$0.881$ &$0.768$      &$0.919$          &$\textbf{0.983}$
                                &$0.916$       &$0.818$   &$0.719$    &$0.871$          &$\textbf{0.952}$
  \\
                                & DIPPER                   &$0.939$       &$0.846$  &$0.767$     &$0.899$          &$\textbf{0.960}$
                                &$0.879$        &$0.783$   &$0.726$    &$0.841$          &$\textbf{0.902}$\\
\midrule

\multirow{3}{*}{OPT-$6.7$}        & No Attack                &$0.995$       &$0.995$   &$\textbf{0.999}$    &$0.993$          &$0.996$      &$0.997$        &$0.997$   &$0.998$    &$0.995$          &$\textbf{0.998}$     \\
                                & GPT-3.5                  &$0.964$       &$0.874$ &$0.769$      &$0.881$          &$\textbf{0.981}$
                                &$0.904$        &$0.815$  &$0.728$     & $0.801$         &$\textbf{0.932}$
         \\
                                & DIPPER                   &$0.936$       &$0.842$ &$0.795$      &$0.883$          &$\textbf{0.964}$
                                &$0.867$        &$0.782$  &$0.743$     &$0.807$          &$\textbf{0.903}$
                    \\ 
\midrule                
\multirow{3}{*}{GPT-J-6B}        & No Attack                &$\textbf{1.000}$       &$0.997$    &$1.000$   &$0.997$          &$0.998$      &$0.995$        &$0.993$    &$\textbf{1.000}$   &$0.995$          &$0.996$     \\
                                & GPT-3.5                  &$0.946$       &$0.874$ &$0.764$      &$0.931$          &$\textbf{0.966}$      &$0.887$        &$0.804$  &$0.721$     &$0.868$         &$\textbf{0.913}$     \\
                                & DIPPER                   &$0.923$       &$0.850$ &$0.803$      &$0.960$          &$\textbf{0.963}$      &$0.855$        &$0.783$   &$0.753$    &$0.898$          &$\textbf{0.901}$  
                    \\

                    \bottomrule
\end{tabular}
\end{table*}

\subsection{Text Quality}
\begin{figure}[ht]
    \centering
    \includegraphics[width=1\linewidth]{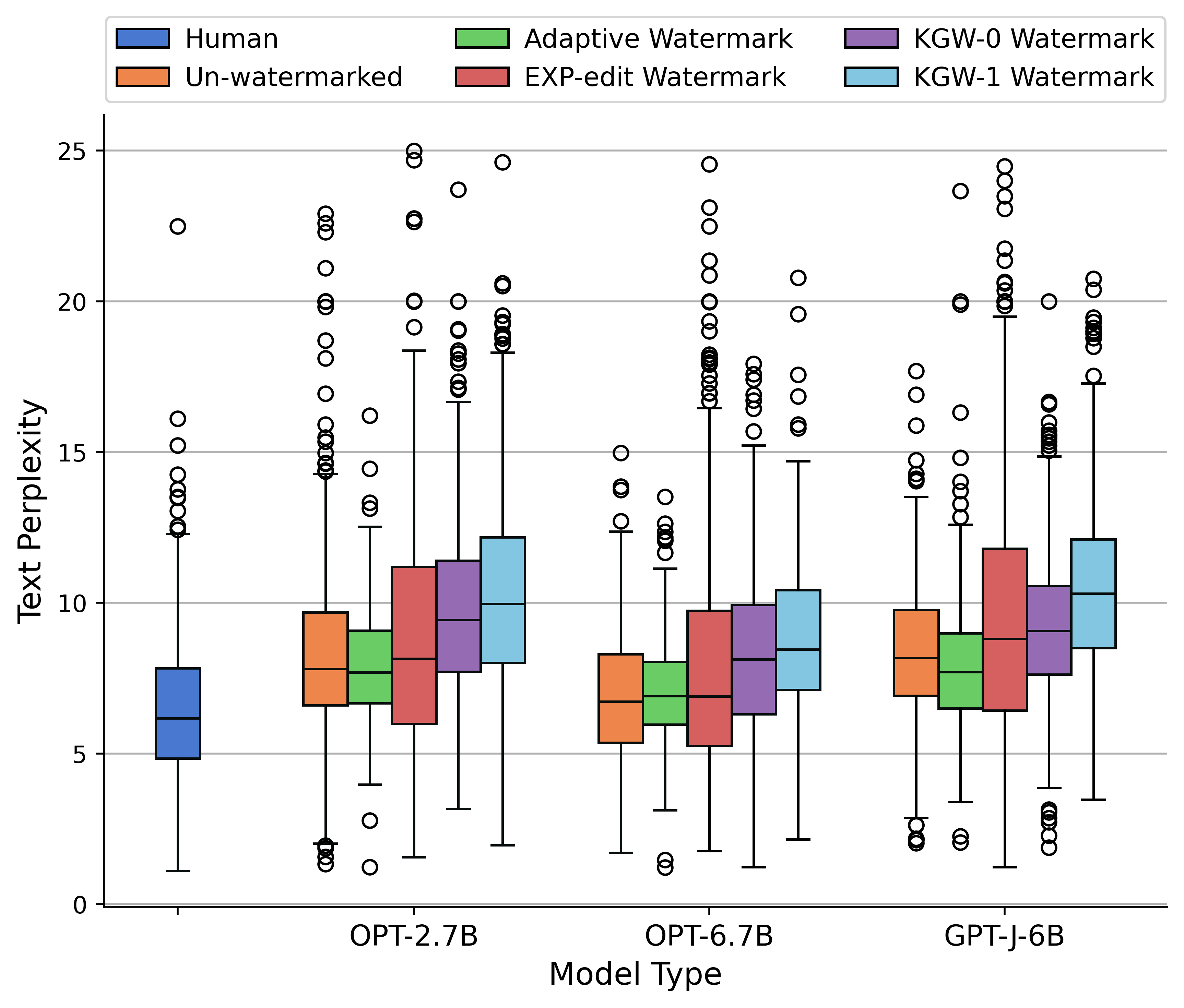}
    % \vspace{-0.7cm}
    \caption{Comparison of text perplexity among human-written text, un-watermarked text, and texts using various watermark methods across different language models. For KGW-0 and KGW-1, the watermark strength and green list size are set as $2.0$ and $0.5$, respectively.}
    \label{fig:pplx}
    % \vspace{-0.5cm}
\end{figure}

The quality of the generated text is evaluated by perplexity. A lower perplexity value suggests better text quality. In Figure \ref{fig:pplx}, we compare the perplexity of text generated by our method with human-written text, un-watermarked text (texts generated by the model without any watermarking), and other baseline methods. Our method has a negligible impact on the perplexity compared to the \emph{un-watermarked} text. It also shows comparable results with EXP-edit, which is a distortion-free watermark method. The quality of text watermarked using our approach is clearly superior to that of KGW-$0$ and KGW-$1$. In Appendix~\ref{app:repetition}, we also report the repetition rate of watermarked text.

\subsection{Security Results}
\label{exp:security}
The spoofing attack \cite{sadasivan2023aigenerated} is used to evaluate the security of our method, which is an adversarial attack designed to produce non-watermarked text that can be mistakenly identified as AI-generated by the watermark detector. This is accomplished by frequently querying the watermarked language model and identifying the potential `green' list as per a particular watermarking algorithm by analyzing token frequencies in the generated text.

Specifically, for KGW-1/KGW-2, the attacker seeks to uncover the green token list for any selected fixed prefix. This entails generating 5,000 sentences under the KGW-1 and examining the frequency of tokens among 181 common tokens ($C$). From $C$, the top 50 tokens with the highest frequency ($H$) are identified. For example, the process involves analyzing the occurrence frequency of tokens following `the.' As the actual green token list ($G$) is determined using the token `the' as a hash, the decryption rate is then computed based on the proportion of tokens in $H$ that belong to $G$.

\begin{table}\scriptsize
\centering
\setlength{\tabcolsep}{3pt}
\caption{Security performance for different watermark methods.}\label{security}
% \vspace{-0.2cm}
\begin{tabular}{clcccc}
\toprule
Language Model           & Evaluation & KGW-0 & KGW-1 & KGW-2 & Ours \\
\midrule 
\multirow{1}{*}{OPT-6.7} & Decryption Rate  &$0.963$                         &$0.912$                         &$0.723$                         &$0.571$                         \\             \bottomrule   
\end{tabular}
% \vspace{-1em}
\end{table}

For the proposed watermarking algorithm, as our `green' token list (positive entries in logits scaling vector $v^{(t)}$) changes with the semantic meaning, the attacker needs to determine the list for any semantic meaning embedding. However, fixing a specific semantic is challenging because the semantics of the generated text can vary with each generation, making it difficult to measure and control consistently. Therefore, to help us qualify the decryption rate, we strengthen the spoofing attack. In our experiments, we maintain a fixed sentence embedding for all generations. This implies that the green token list remains unchanged, allowing us to verify if the green token list can be obtained for this fixed semantic embedding. We analyze the frequency of tokens in a manner similar to the previously mentioned methods. 

We note that the decryption rate derived from this strengthened version of the spoofing attack serves as an upper bound because, in practice, attackers typically only have access to the watermarked LLM API, which generates watermarked text in response to input prompts. Even if an attacker repeatedly inputs the same prompt and generates multiple texts, the generated text may have different semantic meanings, and decrypting the mapping to the green token list becomes significantly more challenging.  

As shown in Table~\ref{security}, we can observe that KGW-0 exhibits the highest decryption rate because of its fixed `green/red' list. For KGW-1 and KGW-2, the implementation of the previous token as a hash key adds a certain level of difficulty to the decryption process compared to the KGW-0. Our method demonstrates superior security performance compared to those works, even for the strengthened spoofing attack, with a decryption rate of only $0.571$.
This is because the AWTI further reduces the fraction of watermarked tokens in the text, thereby diluting the evident token preference within a specific semantic range. More discussion regarding the benefits of AWTI to security can be found in Appendix~\ref{sec:security}.

\subsection{Ablation Study}
In this section, our investigation focuses on assessing the robustness and text quality of watermarked text using our watermark under varying entropy thresholds and watermark strength. Specifically, we adjust the threshold value $\alpha$ to 0, 1, 2, 3 and set the watermark strength $\delta$ at $0.5$, $1$, $1.5$, and $2$. In Appendix~\ref{appendix_additional_ab}, we further analyze the performance of our watermark under different measurement models, and evaluate the effectiveness of each component in our adaptive watermark.

\begin{figure}[ht]
    \centering
    \includegraphics[width=1\linewidth]{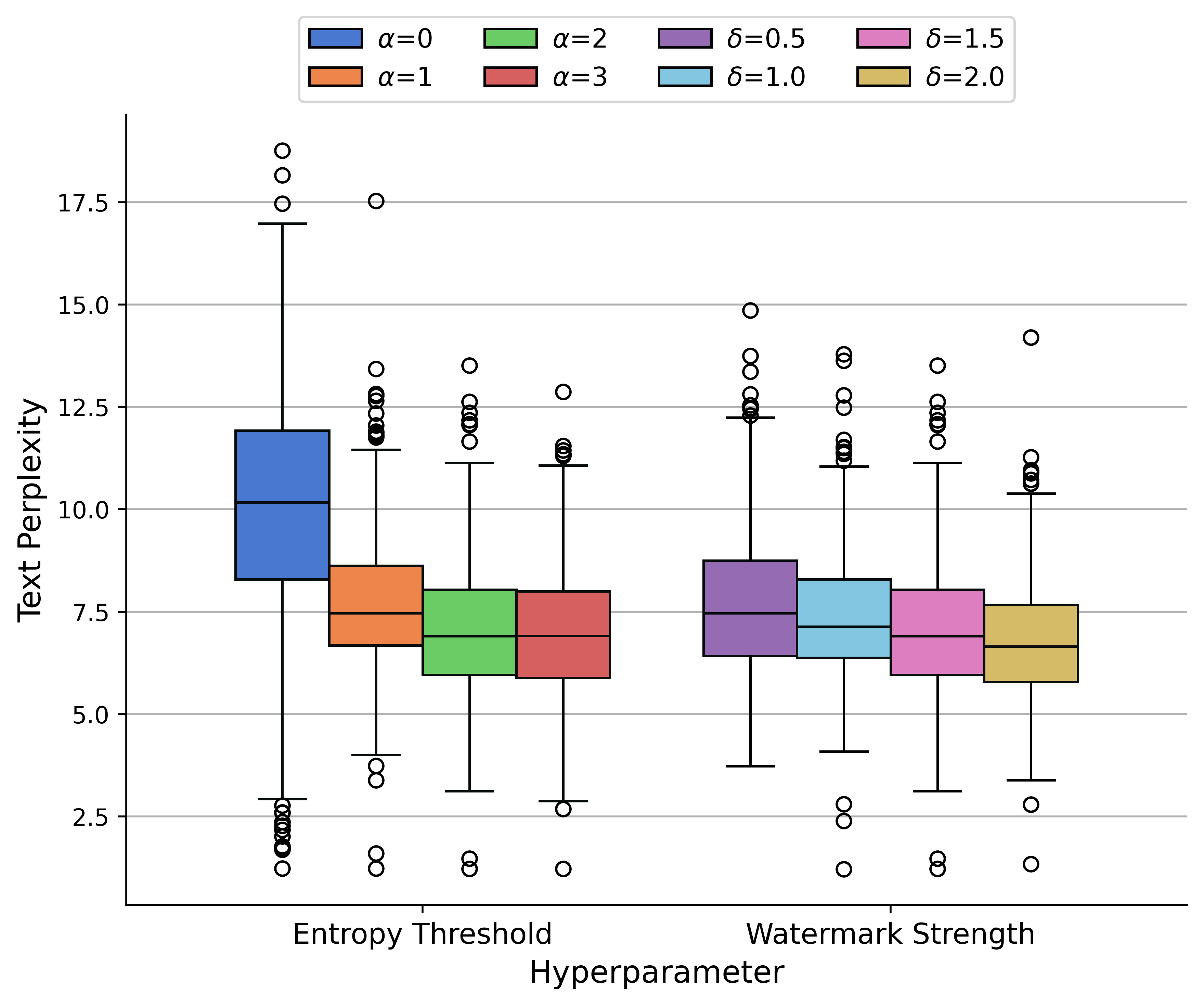}
    % \vspace{-0.7cm}
    \caption{Comparison of text perplexity at varying entropy thresholds (left) and across different watermark strengths (right). $\alpha$ represents the entropy threshold. $\delta$ represents the watermark strength.}
    \label{fig:pplx_ab}
\end{figure}

\textbf{Entropy Thresholds.} In Table \ref{entropy_threshold}, we provide the ROC-AUC, Best F1 Score under different entropy thresholds when the watermarked text is paraphrased using GPT-3.5. In addition, we provide the Average Watermark Rate (AWR), which is the fraction of the watermarked tokens. Figure \ref{fig:pplx_ab} (left) shows the text perplexity comparison under different entropy thresholds. When $\alpha=0$, AWTI is skipped in watermark generation. In this case, the ROC-AUC and Best F1 Score are high, but the watermarked text exhibits high perplexity, with a median value exceeding $10$. When setting $\alpha$ to 1, 2, or 3, there is a notable decrease in the perplexity of the watermarked text, as fewer tokens are perturbed, reflected by the decreasing AWR. 
In addition, the robustness of the watermark is improved when AWTI is applied ($\alpha=1, 2$), potentially because some modified tokens are filtered out by AWTI in detection.  
For entropy threshold $\alpha=3$, our method can achieve results comparable to other methods, with only about half of the watermarked tokens.

\begin{table}[ht]\scriptsize
\centering
\setlength{\tabcolsep}{5.5pt}
\caption{Robustness performance for different entropy thresholds.}\label{entropy_threshold}
% \vspace{-0.2cm}
\begin{tabular}{clcccc}
\toprule
Language Model           & Evaluation & $\alpha=0$ & $\alpha=1$ & $\alpha=2$ & $\alpha=3$ \\
\midrule 
\multirow{3}{*}{OPT-6.7} & ROC-AUC  &$0.978$                         &$0.988$                         & $0.981$                        &$0.972$                         \\
                         & Best F1 Score   &$0.935$                       & $0.957$                        &$0.932$                         &$0.917$        \\
                         & AWR   &$1.000$                       & $0.835$                        &$0.707$                         &$0.553$ \\
                         \bottomrule   
\end{tabular}
\end{table}

\textbf{Watermark Strength.} In Table \ref{watermark_strength} and Figure \ref{fig:pplx_ab} (right), we increase the watermark strength $\delta$ from $0.5$ to $2$ to examine its impact on robustness and perplexity. As shown in Table \ref{watermark_strength}, there is little impact on the ROC-AUC values in terms of robustness. Interestingly, in contrast to previous works \cite{kirchenbauer2023watermark, zhao2023provable} where perplexity tends to rise with larger watermark strength, the perplexity of the text watermarked with our method decreases as the $\delta$ increases in a reasonable range. This benefits from the proposed AWTS, which embeds the watermark by adjusting \textit{temperature} as explained in Section \ref{AWTS}. However, this does not imply that a larger $\delta$ consistently results in better text quality. A higher watermark strength can cause the token selection process to resemble greedy sampling more closely, potentially leading to the generation of repetitive content.

\begin{table}[ht]\scriptsize
\centering
\setlength{\tabcolsep}{5pt}
\caption{Robustness performance for different watermark strength.}\label{watermark_strength}
% \vspace{-0.2cm}
\begin{tabular}{clcccc}
\toprule
Language Model           & Evaluation & $\delta=0.5$ & $\delta=1$ & $\delta=1.5$ & $\delta=2$ \\
\midrule 
\multirow{2}{*}{OPT-6.7} & ROC-AUC  &$0.982$                         &$0.984$                         &$0.981$                         &$0.978$                         \\
                         & Best F1 Score   &$0.938$                       &$0.945$                         &$0.932$                        &$0.938$        \\             \bottomrule   
% \vspace{-0.2cm}
\end{tabular}
\end{table}

\section{Conclusion}
In this paper, we propose a holistic LLM watermark method to achieve the following properties: strong robustness, high security, high text quality, and accurate detection without the need to access the original language model and prompt. Our experiments show that, compared to baseline methods, our approach demonstrates better performance under different paraphrase attacks. Notably, our method has a negligible impact on text perplexity when compared to \emph{un-watermarked} text. Additionally, we highlight the security of our approach by employing the spoofing attack. Specifically, the capacity of language models differs among various models. Therefore, the model owner may need to tune the entropy threshold and watermark strength carefully based on the performance of the language model to get better watermarking performance. Moreover, it is recommended for watermarked model owners to train their own small-sized measurement model based on the watermarked model for performance and security considerations. In future research, we aim to conduct a theoretical analysis to deepen the understanding of the performance characteristics of the watermarking approach and offer an optimal solution. 

\clearpage
\section*{Impact Statement}
Adding a watermark to the LLMs can significantly enhance transparency and accountability by making it easier to identify AI-generated content. Its impact extends broadly in the era of AI, including safeguarding intellectual property (IP), reducing the dissemination of misinformation, and mitigating the misuse of AI-generated content across educational and other domains. Furthermore, looking ahead to societal consequences, effective watermarking has the potential to bolster public trust in AI technologies by offering a clear means of distinguishing between human and AI-generated content.

\bibliography{example_paper}
\bibliographystyle{icml2024}

%%%%%%%%%%%%%%%%%%%%%%%%%%%%%%%%%%%%%%%%%%%%%%%%%%%%%%%%%%%%%%%%%%%%%%%%%%%%%%%
%%%%%%%%%%%%%%%%%%%%%%%%%%%%%%%%%%%%%%%%%%%%%%%%%%%%%%%%%%%%%%%%%%%%%%%%%%%%%%%
% APPENDIX
%%%%%%%%%%%%%%%%%%%%%%%%%%%%%%%%%%%%%%%%%%%%%%%%%%%%%%%%%%%%%%%%%%%%%%%%%%%%%%%
%%%%%%%%%%%%%%%%%%%%%%%%%%%%%%%%%%%%%%%%%%%%%%%%%%%%%%%%%%%%%%%%%%%%%%%%%%%%%%%
\newpage
\appendix
\onecolumn

% \section{Additional Experiment Results}

\section{Semantic Mapping Model Implementation Details}\label{app:exp}
The Semantic Mapping Model is a neural network that starts and ends with linear layers and contains two residual blocks in the middle. For the loss function provided in Section~\ref{SLSVE}, the weight of each term is set as 1.
We train the model with the standard stochastic gradient descent with batch size 128. 

As discussed in Section~\ref{SLSVE}
, we rescale Euclidean distance between two sentence embeddings $d=D(u(S), u(S'))$ to a wider range.
The original range $d\in[L, U]$, where $L=0$ and $U$ is the maximum distance. To enlarge this range, we apply a linear transformation $(T:\mathbb{R}\rightarrow\mathbb{R})$ to the original distance $d$, resulting in a new distance $d'=T(d)\in[L',U']$ where $L'<L$ and $U'>U$. The linear transformation is expressed as
\begin{align}
    \label{eq:linear}
    T(d) = \left(\frac{d-L}{U-L}\right)\cdot(U'-L')+L'.
\end{align}
In our experiment, we set the $L'=-2$ and $U'=4$.

\section{Additional Evaluation Results}\label{appx:additional_exps}

\subsection{Experiment Results on Additional LLM}
\label{app:more_model}
\begin{figure}[h!]
    \centering
    \includegraphics[width=0.5\linewidth]{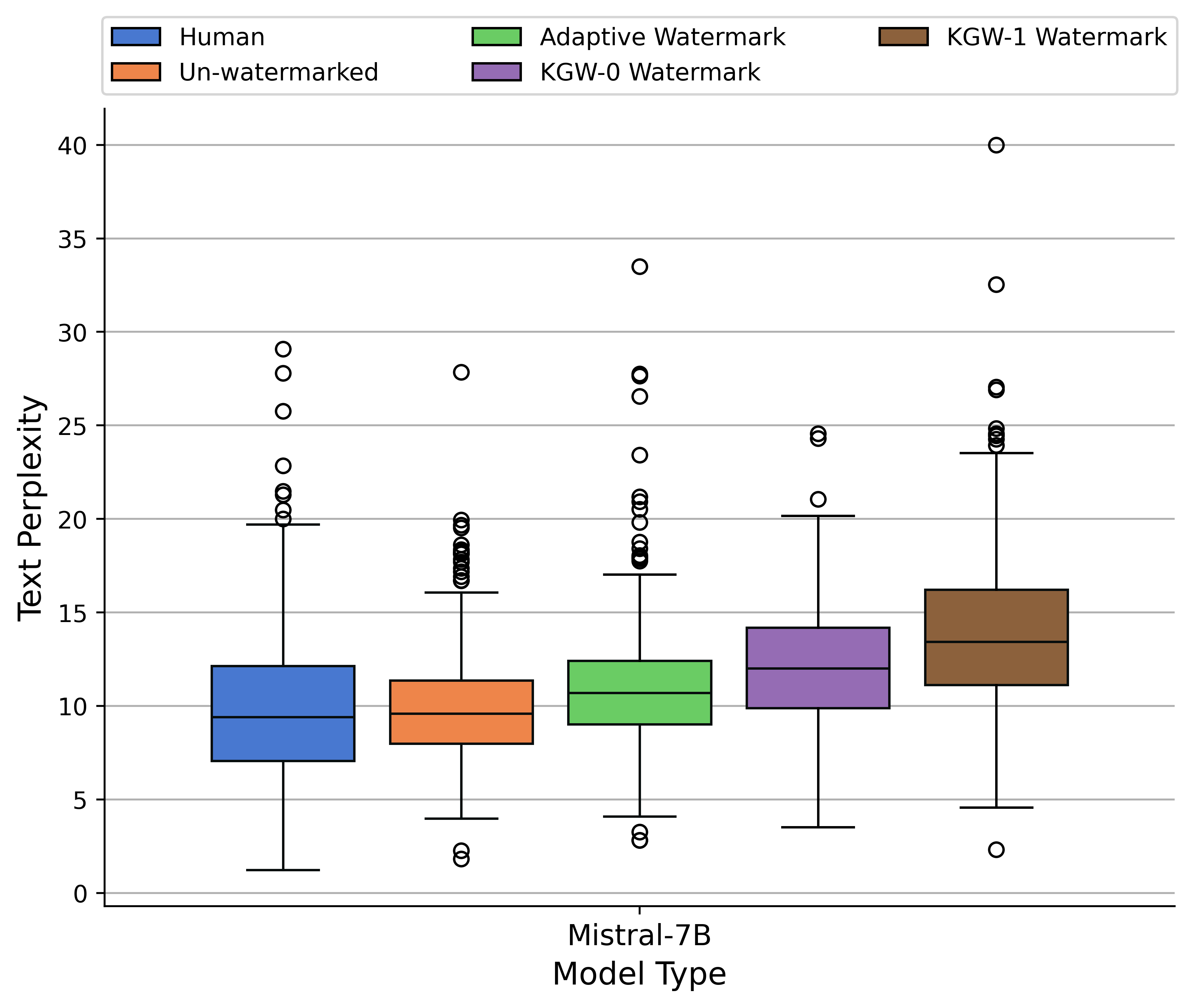}
    \caption{The comparison of text perplexity across various watermarking methods, conducted on Mistral-7B with C4 dataset, with the
perplexity calculated using GPT-3.}
    \label{fig:pplx_mistral7b}
\end{figure}

To further showcase the performance of our watermarking algorithm, we conduct experiments on a more powerful LLM, Mistral-7B, using the C4 dataset. Moreover, we use a more powerful LLM, GPT-3, to calculate the perplexity. As presented in Table~\ref{table:mistral7b} and Figure~\ref{fig:pplx_mistral7b}, our watermarking method not only demonstrates superior robustness performance but also maintains lower text perplexity compared to other methods.

\begin{table}[h!]\scriptsize
\centering
\setlength{\tabcolsep}{3pt}
\caption{The comparison of robustness performance across various watermarking methods, conducted on Mistral-7B with C4 dataset}\label{table:mistral7b}
\vspace{-0.2cm}
\begin{tabular}{clccccc}
\toprule
Language Model           & Evaluation & KGW-0 & KGW-1 &KGW-2 & EXP-edit &  \textbf{Ours} \\
\midrule 
\multirow{2}{*}{Mistral-7B} & ROC-AUC  &$0.952$       &$0.854$ &$	0.792$      &$0.872$          &$0.961$                                              \\
                         & Best F1 Score   &$0.887$        &$0.785$  &$0.747$     & $0.833$         &$0.902$                         \\ 
                         \bottomrule   
\end{tabular}
\end{table}

\subsection{Experiment Results on Additional Dataset}
\label{app:more_data} 
\begin{figure}[h!]
    \centering
    \includegraphics[width=0.5\linewidth]{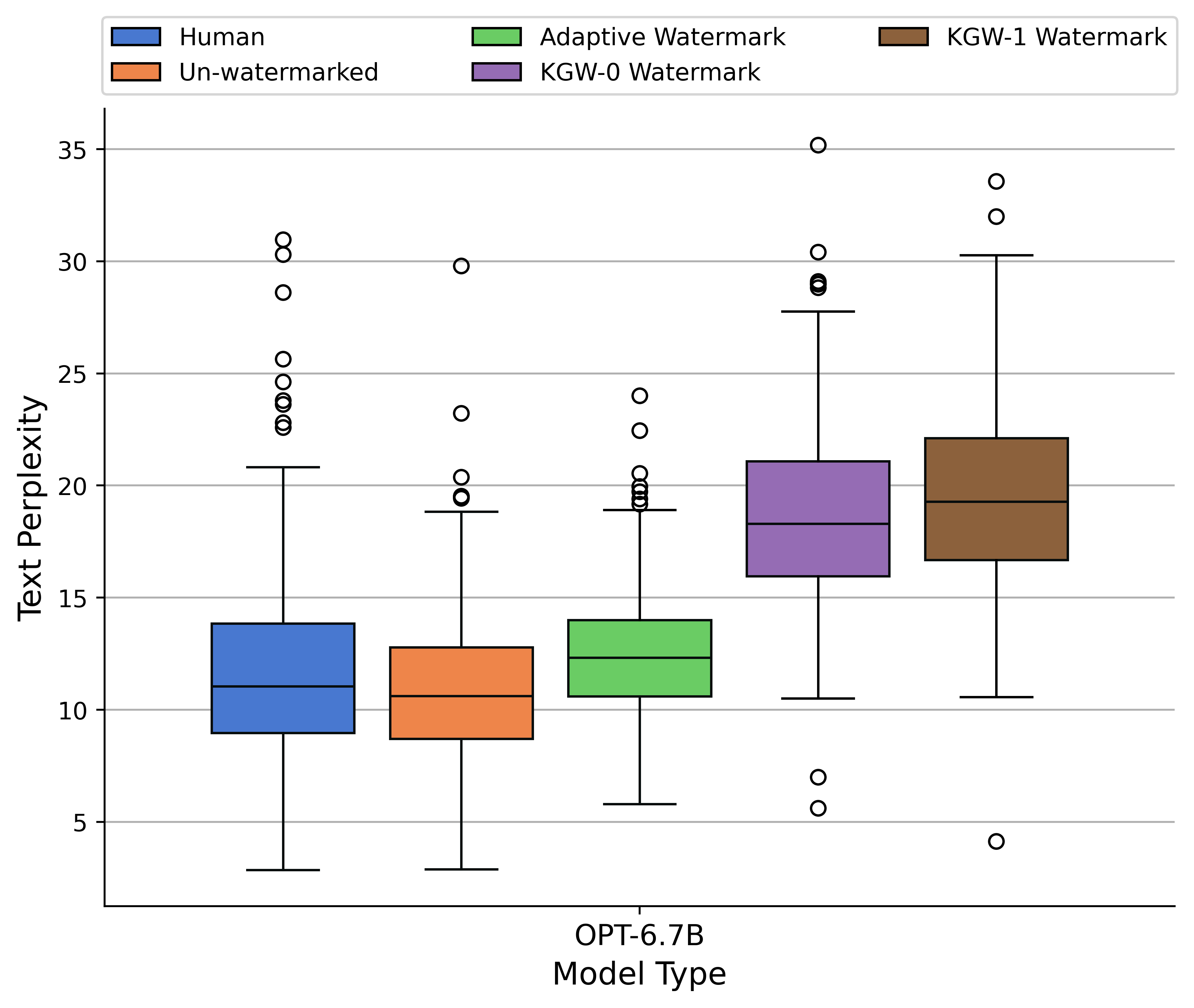}
    \caption{The comparison of text perplexity across various watermarking methods, conducted on OPT-6.7 with ELI5 dataset, with the perplexity calculated using GPT-3.}
    \label{fig:pplx_eli5}
\end{figure}

Besides utilizing C4, which generates subsequent sentences from the first two as prompts, we also consider the long-form question-answering dataset ELI5 to reflect real-world LLM use cases. We conduct experiments with the ELI5 dataset using the OPT-6.7B model to assess robustness performance and text perplexity. In these experiments, watermarked texts were paraphrased by GPT-3, and text perplexity was also calculated using GPT-3. Table~\ref{table:eli5} indicates that our robustness performance is on par with KGW-0, while Figure~\ref{fig:pplx_eli5} demonstrates that the text perplexity associated with our watermarking technique is significantly lower compared to other methods. 

\begin{table}[h!]\scriptsize
\centering
\setlength{\tabcolsep}{3pt}
\caption{The comparison of robustness performance across various watermarking methods, conducted on OPT-6.7 with ELI5 dataset.}\label{table:eli5}
\vspace{-0.2cm}
\begin{tabular}{clccccc}
\toprule
Language Model           & Evaluation & KGW-0 & KGW-1 &KGW-2 & EXP-edit &  \textbf{Ours} \\
\midrule 
\multirow{2}{*}{OPT-6.7} & ROC-AUC  &$0.971$       &$0.781$ &$0.642$      &$0.886$          &$0.967$                                              \\
                         & Best F1 Score   &$0.913$        &$0.733$  &$0.668$     & $0.857$         &$0.899$                         \\ 
                         \bottomrule   
\end{tabular}
\end{table}

\subsection{ROC Curves}\label{appendix_additional_roc}

\begin{figure}[!h]
    \centering
    \subfigure[Paraphrasing attack using GPT-3.5.
]{
        \begin{minipage}[]{1\linewidth}
            \includegraphics[width=1\linewidth]{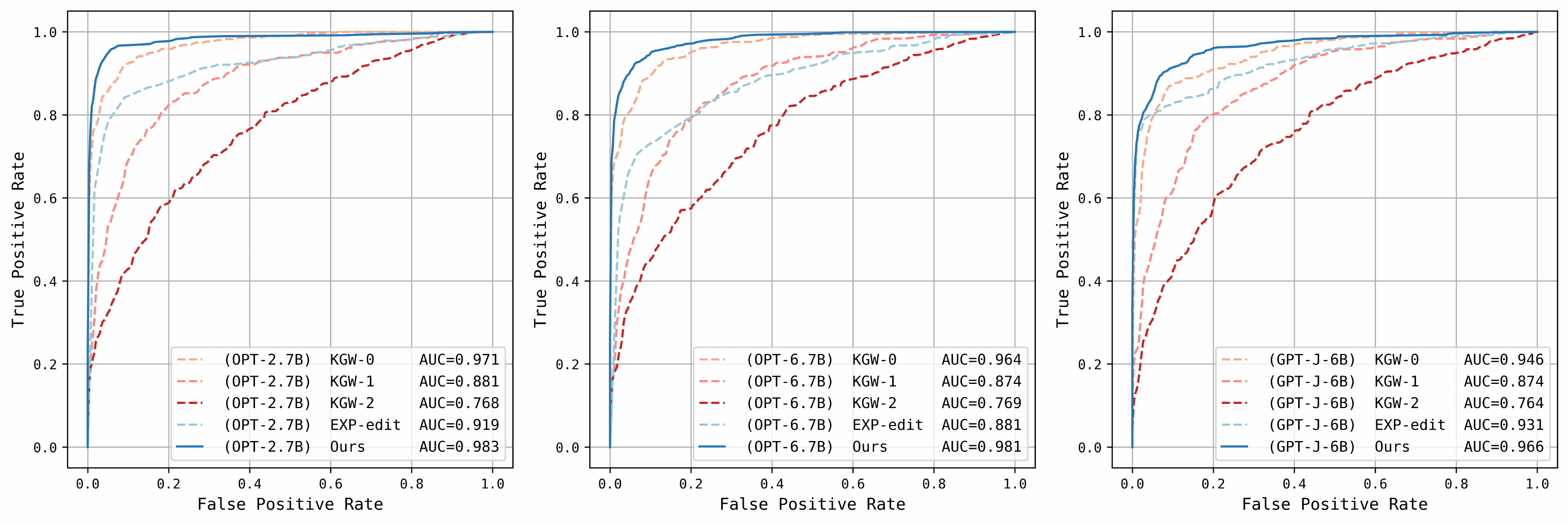}
            \label{fig:rocauc_gptparaphrase}
        \end{minipage}
    }
    \subfigure[Paraphrasing attack using DIPPER.]{
        \begin{minipage}[]{1\linewidth}
            \includegraphics[width=1\linewidth]{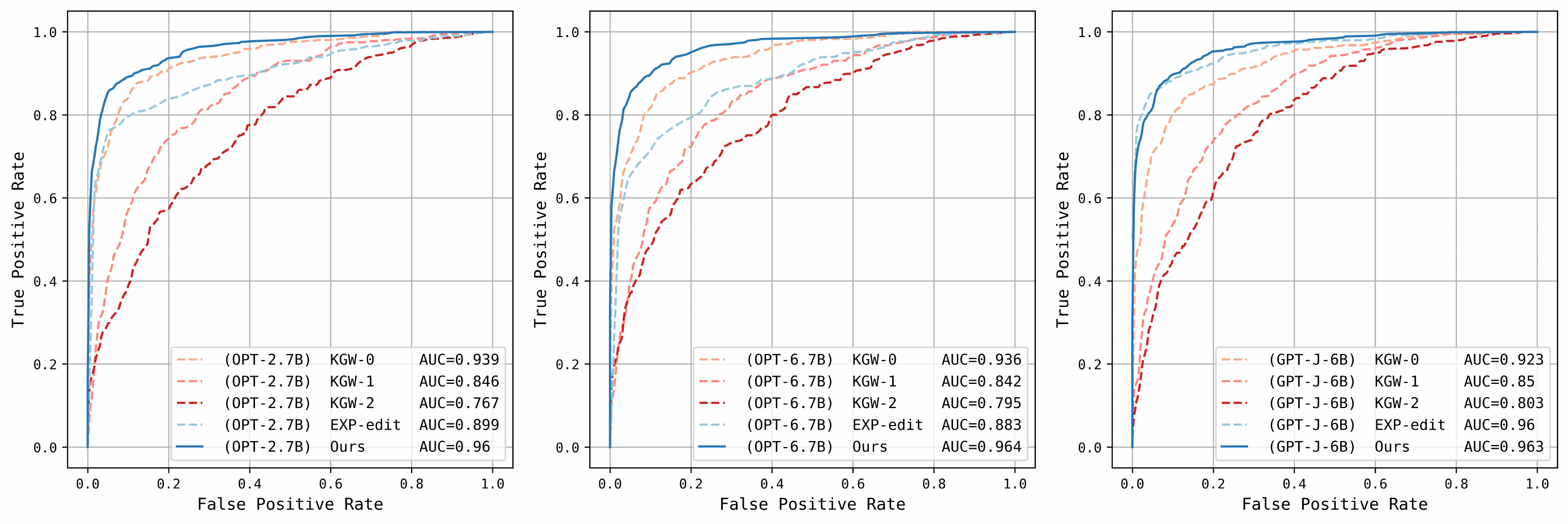}
            \label{fig:rocauc_dipperparaphrase}
        \end{minipage}
    }
    \caption{Comparisons of ROC curves of different watermark methods applied to various language models against two different paraphrasing attacks.}
    \label{fig:roccurve_robust}
\end{figure}
Figure~\ref{fig:roccurve_robust} illustrates the ROC curves and the corresponding AUC values when text is watermarked by different watermarking methods under paraphrasing attack. It is evident that our watermark method exhibits better robustness against paraphrase attacks compared to the other watermarking methods.

\subsection{TPR at Fixed Low FPR}
\label{sub:lowfpr}
In practical applications, it is essential to monitor the TPR at a consistently low FPR to guarantee that un-watermarked texts are not mistakenly identified as watermarked. We showcase the TPR scores at fixed FPR thresholds of 1\% and 10\%, respectively, to demonstrate robustness performance under OPT-6.7B, as depicted in the Table~\ref{table:low_fpr}.

\begin{table}[h]\scriptsize
\centering
\setlength{\tabcolsep}{3pt}
\caption{Robustness performance at fixed low FPR.}\label{table:low_fpr}
\vspace{-0.2cm}
\begin{tabular}{clcccc}
\toprule
Language Model    &Methods       & TPR@1\%FPR & TPR@1\%FPR & ROC-AUC &Best F1 Score  \\
\midrule 
\multirow{5}{*}{OPT-6.7B} & KGW-0  &$0.670$       &$0.896$ &$0.964$                                &$0.904$          \\
                         & KGW-1  &$0.198$        &$0.653$  &$0.874$      &$0.815$                     \\ 
                         & KGW-2  &$0.179$        &$0.455$  &$0.769$       &$0.728$                    \\ 
                         & EXP-edit  &$0.257$        &$0.731$  &$0.881$     &$0.801$                      \\ 
                         & Ours  &$0.779$        &$0.949$  &$0.981$       &$0.932$                    \\ 
                         \bottomrule   
\end{tabular}
\end{table}

\subsection{Repetition Rate}\label{app:repetition}
To examine the repetition problem of watermarked text, we compute the different N-gram repetition rates for the watermarked text generated by different watermarking methods, setting N to 1, 2, and 3 separately. Table~\ref{table:repetition} shows that the repetition rate of our method is similar to that of the EXP-edit, which is a distortion-free watermarking technique.

\begin{table}[h!]\scriptsize
\centering
\setlength{\tabcolsep}{3pt}
\caption{Repetition rate for watermarked text generated by different watermarking methods.}\label{table:repetition}
\vspace{-0.2cm}
\begin{tabular}{clcccc}
\toprule
Language Model    &Methods      & 1-gram & 2-gram & 3-gram & 4-gram  \\
\midrule 
\multirow{5}{*}{OPT-6.7B} & KGW-0  &$0.219$       &$0.046$ &$0.019$                                &$0.012$          \\
                         & KGW-1  &$0.199$        &$0.037$  &$0.013$      &$0.007$                     \\ 
                         & KGW-2  &$0.199$        &$0.036$  &$0.015$       &$0.010$                    \\ 
                         & EXP-edit  &$0.239$        &$0.062$  &$0.026$     &$0.018$                      \\ 
                         & Ours  &$0.241$        &$0.066$  &$0.026$       &$0.014$                    \\ 
                         \bottomrule   
\end{tabular}
\end{table}

% \begin{table}[]\scriptsize
% \centering
% \renewcommand\arraystretch{1.5}
% \begin{tabular}{ccccc}
% \toprule
% Watermarking Method   & Settings                & OPT-2.7B & OPT-6.7B & GPT-J-6B \\
% \midrule
% \multirow{2}{*}{Ours} & GPT2-large              & $29.14$         & $33.16$         &          \\
% % \cmidrule{2-5}
%                       & OPT-125M                & $27.28$         & $31.15$         &          \\
% % \midrule
% KGW-0                & -       &$78.78$          & $81.64$         & $81.06$         \\
% % \midrule
% KGW-1              & -       & $79.15$         & $85.15$         & $84.60$         \\
% % \midrule
% EXP-edit              & -          &$22.12$          &$25.89$          & $25.72$        \\
% \bottomrule
% \end{tabular}
% \end{table}

\section{Comparison with Additional Baselines}
\label{app:sweet_sir}

We note that \citet{lee2023wrote} also discuss the strategy of perturbing only the high-entropy distributions during generation. However, their method requires access to both the original LLMs and prompts. Accessing the original prompt during the detection phase can be impractical, and employing the source watermarked LLM for detection is not only time-consuming but also incurs substantial computational costs.

Moreover, \citet{liu2023semantic} propose a semantic-based watermarking method, which also relies on the original prompts for detection. While this method enhances the security of the watermark, there remains a security risk, as attackers might fix the semantics of prompts to access the corresponding green token list. Our algorithm has a more stable and robust Semantic Mapping Model to handle the perturbation of semantics after modification (e.g., paraphrase), further improving our watermark's robustness. Moreover, our proposed AWTI further strengthens the security and improves the quality of watermarked text. 

\begin{figure}[h!]
    \centering
    \includegraphics[width=0.5\linewidth]{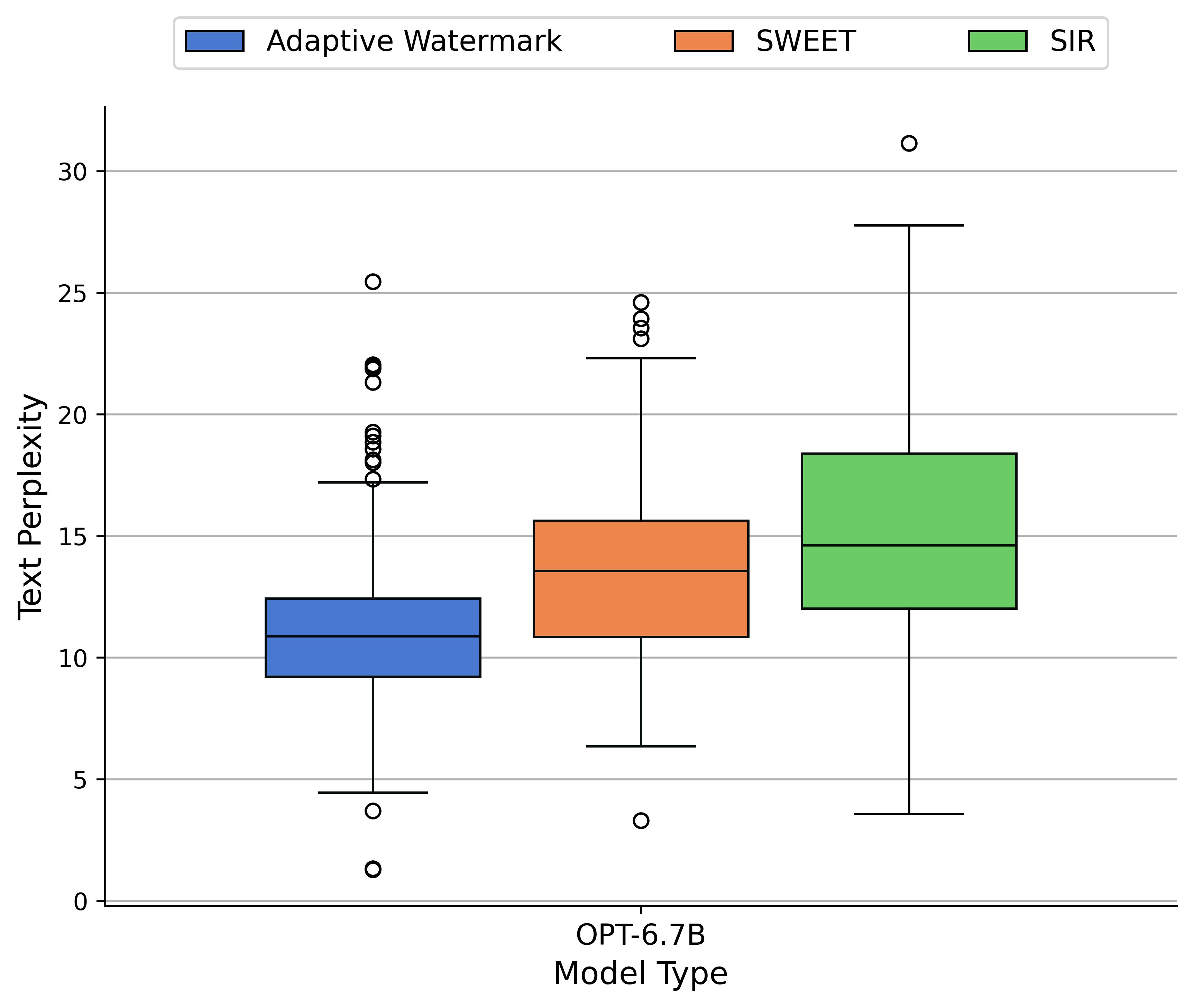}
    \caption{The comparison of text perplexity among Adaptive Watermark, SWEET, and SIR methods, conducted on OPT-6.7, with the perplexity being calculated using GPT-3.}
    \label{fig:pplx_sweet_sir}
\end{figure}

We conduct experiments to evaluate the robustness performance (after paraphrasing by GPT-3) and the perplexity of watermarked text using our algorithm (calculated by GPT-3), comparing it against both SWEET~\cite{lee2023wrote} and SIR~\cite{liu2023semantic}. All experiments are based on the OPT-6.7B model. The results of robustness performance are presented in Table~\ref{table:sweer_sir}, and the results of text perplexity are presented in Figure~\ref{fig:pplx_sweet_sir}. It is worth noting that the comparison is not completely fair, as both SWEET and SIR are not model-agnostic. This implies that their detection processes involve using either the original prompt or the source watermarked LLM. The results show that we achieve better robustness performance and lower perplexity.

\begin{table}[h!]\scriptsize
\centering
\renewcommand\arraystretch{1.1}
\caption{Robustness performance comparison among Adaptive Watermark, SWEET, and SIR methods.}\label{table:sweer_sir}
\vspace{-0.2cm}
\begin{tabular}{clcc}
\toprule
Language Model   & Methods   & ROC-AUC & Best F1 Score\\
\midrule
 \multirow{3}{*}{OPT-6.7}        & SWEET                 & $0.847$         & $0.787$         \\
              & SIR                & $0.963	$         & $0.903$         \\
             & Ours                  &$0.981$          & $0.932$        \\
\bottomrule
\end{tabular}
\end{table}

\section{Security Discussion}
\label{sec:security}

The security of a watermarking algorithm here refers to the difficulty an attacker has in statistically analyzing and decrypting the watermarking rules using the watermarked texts with the aim of forging the watermarks. For previous approaches such as KGW-$k$, the rule of green/red list can be simply recovered by analyzing token frequency after a specific fixed token, as the green/red list is purely determined using $k$ previous tokens. For our semantic mapping model, the input sentence embedding space is continuous, and it depends on all the previously generated tokens, which requires the attackers to analyze token frequency based on fixed semantics. However, it is hard in practice, as even with the same prompt, the semantics of the generated text could vary widely. Consequently, for attackers aiming to identify the green token list associated with a particular semantic meaning, more effort or considerably larger sample size is required for our algorithm to conduct a meaningful analysis.

In general, we believe that a watermarking algorithm with more complex procedures will necessitate greater effort or a larger dataset for attackers to forge watermarking. To see this, we consider the examples of KGW-0 and KGW-$k$. KGW-0 employs a globally fixed green/red token list for generating watermarked text, making it the most straightforward to decrypt, i.e., simply analyzing overall token frequency. In contrast, KGW-$k$ enhances security by utilizing previously generated tokens to formulate the green/red token list, requiring attackers to analyze token frequency in relation to specific tokens, thereby complicating the decryption process. Our watermarking algorithm further advances the process by deriving the green/red token list from the semantics of previously generated text and adaptively applying it to tokens with high entropy. As the attackers cannot accurately identify watermarked tokens (high entropy tokens measured using the specific measurement model), this adds another layer of difficulty in statistically analyzing the token frequency, even within a specific semantic context. Therefore, the complexity of our watermarking algorithm makes it more challenging for attackers to decrypt the watermarking rules, thereby enhancing the security of the watermarking algorithm.

Specifically, the implementation of AWTI results in fewer green tokens (positive entries in logits scaling vector $v^{(t)}$) within the generated text compared to other methods like KGW-0/KGW-1/KGW-2. However, the power of our watermark will not be affected. As outlined in Section 4 of our detection process, we prioritize filtering out potential un-watermarked tokens ($U$) and focus on the proportion of green tokens ($\hat{G}$) among potential watermarked tokens ($W$), rather than the proportion of green tokens ($G$) among all tokens ($U \cup W$) in the generated text, as done in KGW-0/KGW-1/KGW-2. Consequently, while the overall ratio of green tokens to all tokens in the generated text ($\frac{|G|}{|U \cup W|}$) may drop, the proportion of green tokens among potential watermarked tokens ($\frac{|\hat{G}|}{|W|}$) might not necessarily decrease. To validate this, we analyzed paraphrased watermarked texts, calculating the average percentage of green tokens across 500 texts generated by KGW-0 and comparing it with the average percentage of green tokens ($\frac{|\hat{G}|}{|W|}$) across 500 texts produced by our watermarking algorithm. The findings, detailed in Table~\ref{table:green_percentage}, reveal that although the overall percentage of green tokens in the generated text ($\frac{|G|}{|U \cup W|}$) stands at 0.499, the percentage of green tokens relative to potential watermarked tokens ($\frac{|\hat{G}|}{|W|}$) reaches 0.705, surpassing that of KGW-0. Furthermore, since the overall percentage of green tokens in the generated text is only about 0.5, this indicates that our generated texts do not exhibit a noticeable bias towards a specific group of tokens. This characteristic makes it more challenging for attackers to decrypt the watermarked texts by analyzing token frequency.

\begin{table}\scriptsize
\centering
\renewcommand\arraystretch{1.1}
\caption{Green token percentage for different methods.}\label{table:green_percentage}
\vspace{-0.2cm}
\begin{tabular}{ccc}
\toprule
   Methods   & Green token percentage & ROC-AUC\\
\midrule
       KGW-0                 & $0.649$         & $0.964$         \\
              Ours ($\frac{|G|}{|U \cup W|}$)	                & $0.499$         & $-$         \\
             Ours ($\frac{|\hat{G}|}{|W|}$)                 &$0.706$          & $0.981$        \\
\bottomrule
\end{tabular}
\end{table}

\section{Additional Ablation Study}
\label{appendix_additional_ab}
\textbf{Measurement Models.} We investigate how different Measurement Models (MMs) impact the robustness of our watermarks and the quality of the generated text. For this purpose, we employ MMs with different sizes and types, such as GPT2 (124M), GPT2-large (774M), OPT-125M, and OPT-350M, in our evaluation. For all models, we set the $\alpha$ and $\delta$ as $2$ and $1.5$, respectively. As shown in Table~\ref{table:ab_measurement_model}, there is only a minor influence on the robustness of the watermark when different MMs are adopted. 

\begin{table}[h!]\scriptsize
\centering
\setlength{\tabcolsep}{3pt}
\caption{Robustness performance for different Measurement Models.}\label{table:ab_measurement_model}
\vspace{-0.2cm}
\begin{tabular}{clcccc}
\toprule
Language Model           & Evaluation & GPT2 & GPT2-large &OPT-125M & OPT-350M  \\
\midrule 
\multirow{3}{*}{OPT-6.7} & ROC-AUC  &$0.985$                         &$0.981$            &$0.987$             &$0.982$                                                \\
                         & Best F1 Score   &$0.951$                         &$0.932$    &$0.947$                     &$0.938$                               \\ 
                         & AWR &$0.778$      &$0.707$      &$0.755$      &$0.718$ \\
                         \bottomrule   
\end{tabular}
\end{table}

\begin{figure}[h!]
    \centering
    \includegraphics[width=0.5\linewidth]{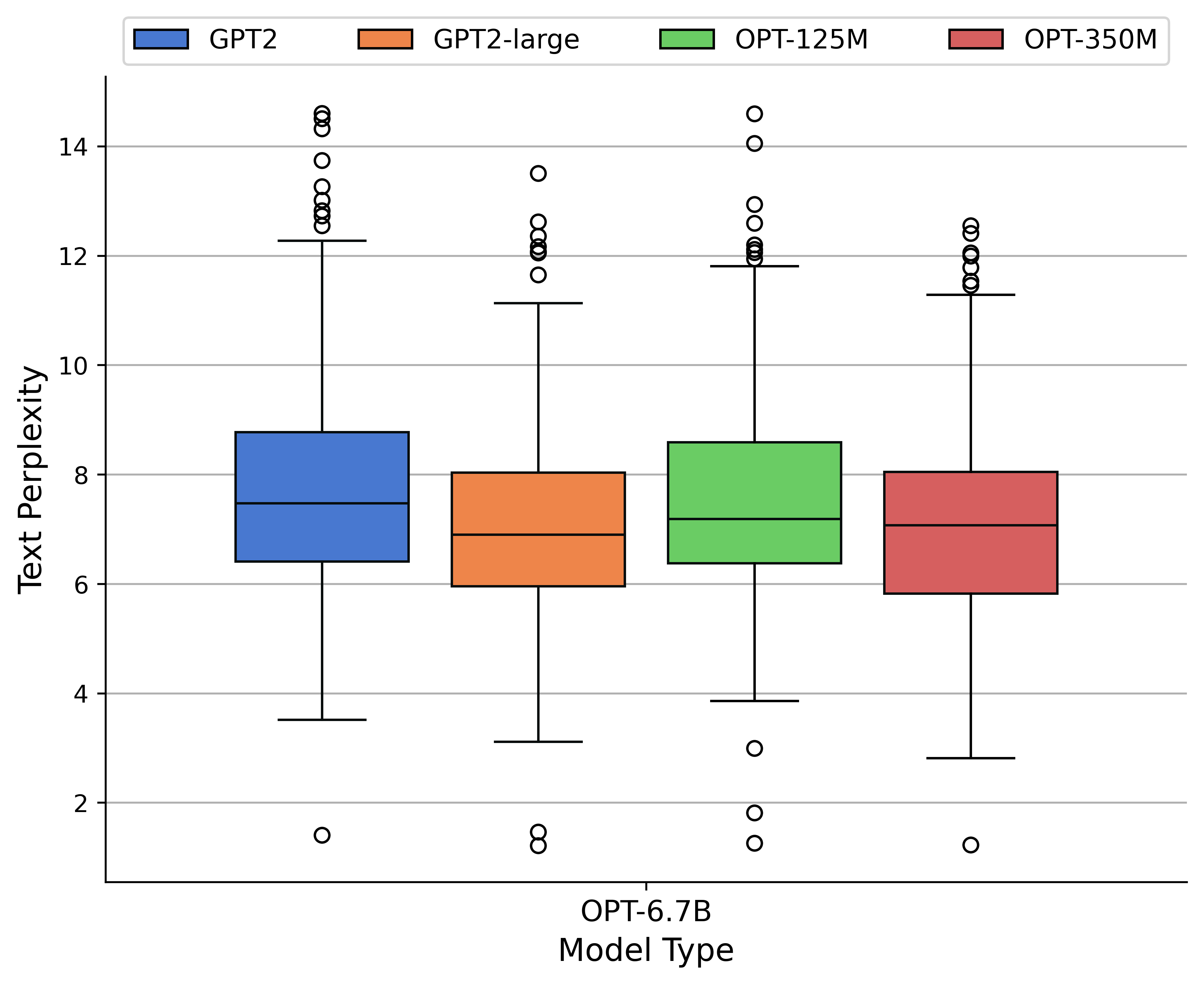}
    \caption{Text perplexity comparison between different Measurement Models. }
    \label{fig:pplx_mm_ab}
\end{figure}

However, the effect on text quality is more nuanced. 
As indicated in Figure~\ref{fig:pplx_mm_ab}, models like GPT2-large and OPT-350M will result in lower perplexity compared to the GPT2 and OPT-125M. One primary reason is that different MMs produce different entropy estimates, which results in different AWRs even with the same entropy threshold. Generally, a higher AWR implies a larger number of watermarked tokens, which can slightly impact text quality.

\begin{figure}[h!]
    \centering
    \includegraphics[width=0.5\linewidth]{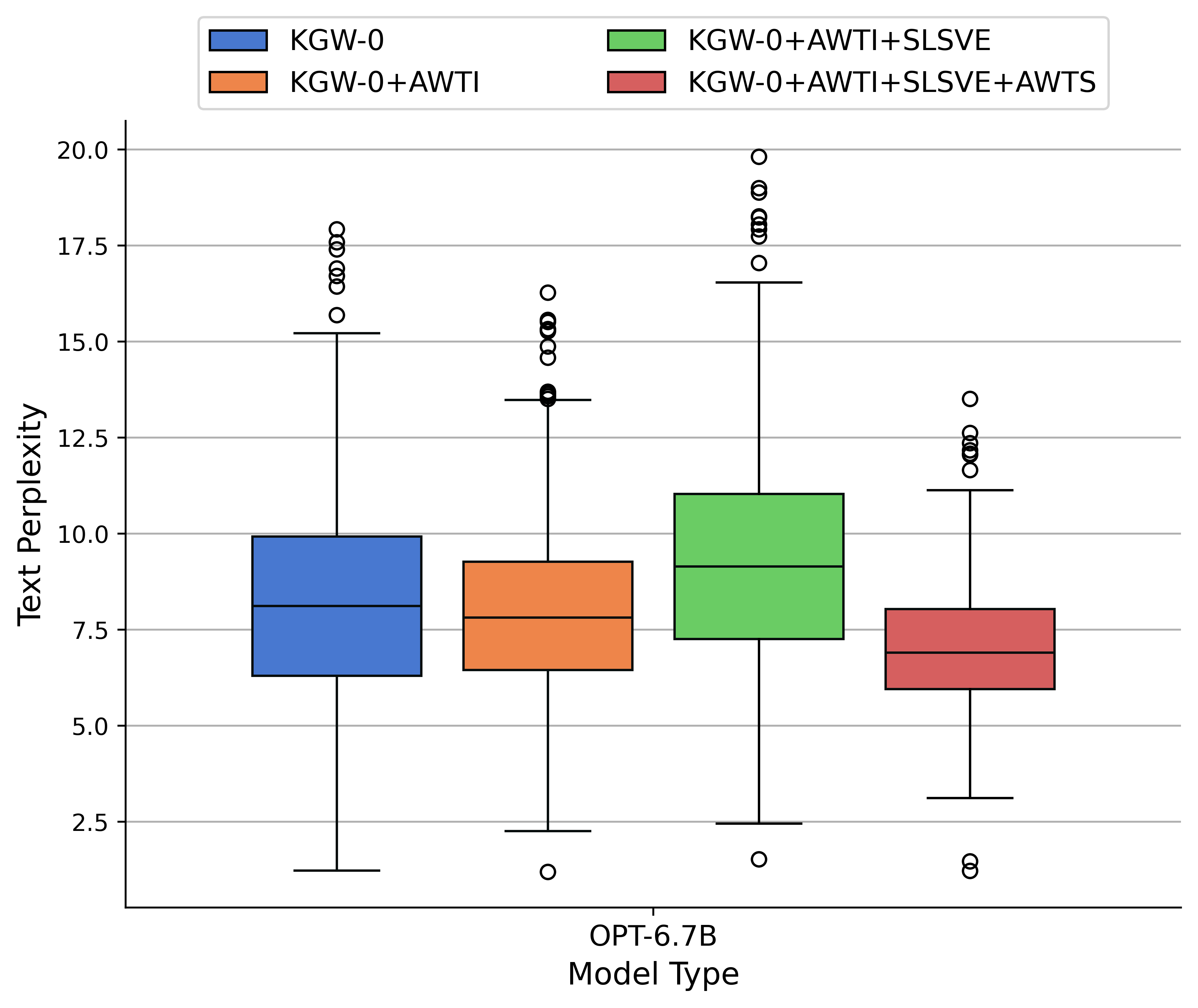}
    \caption{Text perplexity comparison after the removal of different components. }
    \label{fig:pplx_ab_components}
\end{figure}

\textbf{Effectiveness of Each Component in Adaptive Watermark.} Our proposed watermarking method mainly consists of three key components:  Adaptive Watermark Token Identification (AWTI), Semantic-based Logits Scaling Vector Extraction (SLSVE), and Adaptive Watermark Temperature Scaling (AWTS). Here, we use KGW-0 as our baseline model and analyze the robustness and perplexity of watermarked text by sequentially adding our three components. Table~\ref{table:components_ab} and Figure~\ref{fig:pplx_ab_components} exhibit the robustness and text perplexity performance under the following four different settings. \texttt{KGW-0+AWTI} represents applying AWTI to the KGW-0, aiming to adaptively perturb the distributions with low entropy. \texttt{KGW-0+AWTI+SLSVE} indicates that both AWTI and SLSVE are applied to KGW-0. \texttt{KGW-0+AWTI+SLSVE+AWTS} represents our complete Adaptive Watermark, where all our three components are applied. 

\begin{table}[h!]\scriptsize
\centering
\renewcommand\arraystretch{1.1}
\caption{Robustness performance comparison after the removal of different components.
}\label{table:components_ab}
\vspace{-0.2cm}
\begin{tabular}{clcc}
\toprule
Language Model   & Settings   & ROC-AUC & Best F1 Score\\
\midrule
 \multirow{4}{*}{OPT-6.7}        & KGW-0                 & $0.964$         & $0.904$         \\
              & KGW-0 + AWTI                & $0.962$         & $0.904$         \\
             & KGW-0 + AWTI + SLSVE                  &$0.960$          & $0.893$        \\
 & KGW-0 + AWTI + SLSVE + AWTS                       & $0.981$         & $0.932$         \\
\bottomrule
\end{tabular}
\end{table}

Table~\ref{table:components_ab} and Figure~\ref{fig:pplx_ab_components} demonstrate that the application of AWTI reduces the perplexity of the watermarked text without compromising robustness performance.
The introduction of SLSVE enhances the security of the watermark but comes at the cost of increased perplexity in the watermarked text.
Finally, when all three components are applied, our watermark method excels in both robustness and perplexity performance compared to existing baselines.

\section{Watermarked Text Generation time}
In this section, we compare the watermarked text generation time and the watermark detection time for different approaches. For each method, we generate and detect 100 instances of watermarked text, each containing roughly 200 tokens. Subsequently, we compute the average time taken for both the generation and detection of each instance of watermarked text. For our method, we set the value of the entropy threshold $\alpha = 2$ and the watermark strength $\delta = 1.5$. We evaluate the generation and detection time of the proposed adaptive watermarking with two different Measurement models.

Table~\ref{table:generation_speed} presents all the results. For generation time, despite employing three additional models in our watermarked text generation process, our average generation time remains faster than that of KGW-0 and KGW-1. Our method is slightly slower than EXP-edit, primarily because EXP-edit uses a sampling-based watermarking approach that does not require logit computation. Specifically, using a Measurement model with smaller parameters can lead to faster generation times. Regarding the time required for detection, our method is slightly slower than KGW-0 and KGW-1. However, it is significantly faster than EXP-edit, which necessitates computing the alignment between the watermarked text and the watermark key sequence.

\begin{table}[h]\scriptsize
\centering
\renewcommand\arraystretch{1.1}
\caption{Watermarked text generation and detection time across various watermarking methods.}\label{table:generation_speed}
\vspace{-0.2cm}
\begin{tabular}{ccccc}
\toprule
Language Model &Watermarking Method   & Settings               &Average Generation Time & Average Detection Time\\
\midrule
% \midrule
\multirow{5}{*}{OPT-6.7} &KGW-0                & -                & $81.64$s         & $0.01$s         \\
% \midrule
&KGW-1              & -                & $85.15$s         & $1.73$s         \\
% \midrule
&EXP-edit              & -                  &$25.89$s          & $163.19$s        \\
&Ours & GPT2-large                       & $33.16$s         & $5.57$s         \\
 &Ours                     & OPT-125M                        & $31.15$s         &$3.63$          \\
\bottomrule
\end{tabular}
\end{table}

% \clearpage

\section{Watermarked Text Example}
We present some examples of watermarked text using our proposed method. Table~\ref{watermarked_text_example} shows the un-watermarked text and watermarked text under different entropy thresholds ($\alpha$). The watermarked texts are generated using the OPT-6.7B model with the watermark strength $\delta=1.5$. The watermarked tokens are highlighted in yellow. As the value of $\alpha$ increases, it is evident that the proportion of watermarked tokens noticeably decreases.

\begin{center}\scriptsize 
    % [!ht] \scriptsize 
    \centering 
    \begin{longtable}{l|p{10cm}}
        \caption{Examples of watermarked text using our method with various entropy thresholds. In these examples, tokens highlighted in yellow represent watermarked tokens, indicating that these tokens were sampled from a perturbed distribution.} 
    \label{watermarked_text_example}  
    \setlength{\tabcolsep}{15pt}  \\
    \toprule
     \multirow{3}{*}{Prompt} &  It's not always easy to keep mentally focused during running. Focus requires discipline and a certain level of mental toughness, meaning that you have developed the capacity to perform to the best of your ability despite external conditions or internal distractions. \\ \midrule
     \multirow{11}{*}{Un-watermarked text} &  While it's not impossible to train for a specific event and keep mentally focused at all times, it's challenging. Here are three things you can do to help yourself.
     
Plan Ahead

The best time to think about race strategy is during warm-ups before any race. Once your legs get moving, they will tell you what they need—you just have to listen to them! It’s also a good idea to write down in detail the training plan for your event and make sure that you understand it completely. Plan your warm-up so that there is an emphasis on building speed without injury risk.

Get Out of Your Head

There are lots of other things to focus your attention on besides running. If you’re running around town, take a walk break every quarter mile or so. This allows you to stretch, cool off and rest, while keeping your muscles engaged.
\\ \midrule
    \multirow{13}{*}{Watermarked text ($\alpha=1$)} & \hl{The}\hl{ ability}\hl{ and}\hl{ desire}\hl{ for}\hl{ discipline}\hl{ are}\hl{ developed}\hl{ through}\hl{ practice}\hl{.}\hl{ The}\hl{ first}\hl{ step}\hl{ in}\hl{ developing}\hl{ discipline}\hl{ during}\hl{ running}\hl{,}\hl{ especially}\hl{ if}\hl{ a}\hl{ runner}\hl{ wants}\hl{ a}\hl{ higher}\hl{ level}\hl{ performance}\hl{,}\hl{ begins}\hl{ with}\hl{ setting}\hl{ goals}\hl{ and}\hl{ working}\hl{ towards}\hl{ meeting}\hl{ these}\hl{ goals}\hl{.}\hl{ }\hl{ }\hl{Goal}\hl{ setting}\hl{ allows}\hl{ runners}\hl{ an}\hl{ avenue}\hl{ to}\hl{ create}\hl{ the}\hl{ mental}\hl{ toughness}\hl{ needed} to\hl{ achieve}\hl{ their}\hl{ best}\hl{ results}\hl{ during}\hl{ training}\hl{ runs}\hl{ or}\hl{ races}.\hl{ Setting}\hl{ realistic}\hl{ expectations}\hl{ will}\hl{ also}\hl{ allow}\hl{ runners} to\hl{ work}\hl{ within}\hl{ their}\hl{ abilities}\hl{ instead} of\hl{ trying} to\hl{ perform}\hl{ beyond}\hl{ what}\hl{ they}\hl{ are}\hl{ currently} capable of.\hl{
} Setting\hl{ realistic}\hl{ expectations}\hl{ can}\hl{ also}\hl{ serve} as\hl{ motivation}\hl{ for}\hl{ runners}\hl{ to}\hl{ continue}\hl{ training}\hl{ despite}\hl{ the}\hl{ obstacles}\hl{ that}\hl{ life}\hl{ has}\hl{ to} offer\hl{ during}\hl{ the}\hl{ course} of\hl{ training}\hl{ or}\hl{ race}\hl{ day}.\hl{ It}\hl{ can}\hl{ also}\hl{ serve} as\hl{ motivation}\hl{ to}\hl{ continue}\hl{ training}\hl{ even}\hl{ when}\hl{ injuries}\hl{ or}\hl{ other}\hl{ problems}\hl{ make}\hl{ training}\hl{ challenging}\hl{ or}\hl{ downright}\hl{ unpleasant}.\hl{
} For\hl{ runners}\hl{ who}\hl{ want}\hl{ to}\hl{ set}\hl{ realistic}\hl{ expectations}\hl{ for}\hl{ themselves}\hl{ during}\hl{ training}\hl{ or}\hl{ race} day,\hl{ the}\hl{ below}\hl{ checklist}\hl{ can}\hl{ serve} as\hl{ guidelines}\hl{:}

Mentally\hl{ prepared}\hl{ -}\hl{ Have}\hl{ enough}\hl{ sleep}\hl{ the} night before\hl{ the}\hl{ race}\hl{ or}\hl{ training}\hl{ runs}\hl{
}

St\hl{ress}\hl{ free} environment\hl{ -}\hl{ Remove}\hl{ negative}\hl{ people}\hl{ from}\hl{ the}\hl{ environment}\hl{
}

Eat\hl{ right}\hl{ -}\hl{ Eat}\hl{ healthy}\hl{ meals}\hl{ the} day before\hl{ the} race or training runs  \\

 \midrule
     \multirow{12}{*}{Watermarked text ($\alpha=2$)} & \hl{The}\hl{ best}\hl{ runners}\hl{ are}\hl{ often}\hl{ described}\hl{ in}\hl{ terms}\hl{ like}\hl{,}\hl{ focused}\hl{,}\hl{ determined}\hl{,}\hl{ and}\hl{ driven}\hl{.}\hl{ You}\hl{ may}\hl{ have}\hl{ read}\hl{ that}\hl{ somewhere}\hl{ on}\hl{ a}\hl{ running}\hl{ magazine}\hl{ cover}\hl{,}\hl{ but}\hl{ did}\hl{ anyone}\hl{ actually}\hl{ write}\hl{ that}\hl{ down}\hl{?}\hl{
}\hl{
}\hl{I}\hl{ was}\hl{ recently}\hl{ asked}\hl{ if}\hl{ I}\hl{ knew}\hl{ any}\hl{ specific}\hl{ exercises} to\hl{ help}\hl{ people}\hl{ stay}\hl{ mentally}\hl{ focused}\hl{ during}\hl{ training} runs.\hl{ My}\hl{ response} was\hl{ "}\hl{No}\hl{,"}\hl{ but}\hl{ the}\hl{ question}\hl{ got} me thinking\hl{ more} about\hl{ how}\hl{ to}\hl{ develop}\hl{ mental}\hl{ toughness}\hl{ for}\hl{ runners}.\hl{
}
St\hl{aying} mentally\hl{ prepared}\hl{ is}\hl{ an}\hl{ important}\hl{ part} of\hl{ the}\hl{ training} process that\hl{ can}\hl{ be}\hl{ just} as important as\hl{ physical}\hl{ preparation},\hl{ but}\hl{ many} runners\hl{ don}'t\hl{ give} it\hl{ much} thought.\hl{
}
It's\hl{ easy} to\hl{ see}\hl{ how}\hl{ physical}\hl{ preparation}\hl{ is}\hl{ important}\hl{ for}\hl{ race}\hl{ day}\hl{ performance}, but\hl{ what} exactly does it mean when\hl{ you}\hl{ talk} about\hl{ mental} preparation?\hl{
}
Here are\hl{ five}\hl{ things} you need to\hl{ do}\hl{ to}\hl{ stay} mentally prepared\hl{ while} running:

1.\hl{ Have}\hl{ realistic} goals and\hl{ expectations}.\hl{
}
Some people\hl{ set}\hl{ unrealistic} expectations\hl{ for} themselves\hl{ or}\hl{ others}\hl{ regarding}\hl{ the}\hl{ number} of miles they\hl{'ll} run\hl{ each} week,\hl{ their}\hl{ times} for\hl{ certain}\hl{ races}, or\hl{ even}\hl{ their}\hl{ overall}\hl{ goal}\hl{ for} their\hl{ training}\hl{ program}.
 \\ \midrule
    \multirow{10}{*}{Watermarked text ($\alpha=3$)} & \hl{The}\hl{ best}\hl{ runners}\hl{ are}\hl{ focused}\hl{ on}\hl{ a}\hl{ specific}\hl{ task}\hl{,}\hl{ which}\hl{ allows}\hl{ for}\hl{ a}\hl{ level}\hl{ playing}\hl{ field}\hl{.}\hl{ The}\hl{ same}\hl{ applies}\hl{ in}\hl{ any}\hl{ aspect}\hl{ in}\hl{ which}\hl{ concentration}\hl{ and}\hl{ discipline}\hl{ are}\hl{ necessary}\hl{.}\hl{
}\hl{
}\hl{In}\hl{ order}\hl{ for}\hl{ a}\hl{ runner}\hl{,}\hl{ cyclist}\hl{,}\hl{ tennis}\hl{ star}\hl{,}\hl{ gol}\hl{ffer}\hl{,}\hl{ etc}. to\hl{ perform} their best, they must\hl{ have}\hl{ the}\hl{ ability} to\hl{ stay} focused at all times.\hl{
}
To\hl{ achieve} this state of focus or\hl{ mental}\hl{ toughness} requires\hl{ training} (\hl{mental} conditioning) and\hl{ practice}.\hl{ You} do not\hl{ have} to\hl{ run}\hl{ every} day for\hl{ hours},\hl{ but}\hl{ just}\hl{ like}\hl{ anything} else it's\hl{ about}\hl{ putting} in the time and\hl{ practicing}\hl{ until} you get better over time.

So\hl{ how} does\hl{ this} apply to\hl{ your}\hl{ life}? The\hl{ best}\hl{ runners}\hl{ know} that\hl{ running}\hl{ is}\hl{ an}\hl{ important} part of their\hl{ lives}, as well as\hl{ their}\hl{ training} routine and\hl{ their}\hl{ overall}\hl{ health}.\hl{
}
The\hl{ best}\hl{ runners}\hl{ also} understand what\hl{ is}\hl{ required} to\hl{ keep} them\hl{ mentally} strong during\hl{ the}\hl{ race}.\hl{ They} know how to\hl{ train}\hl{ themselves} so that they will always be ready and\hl{ prepared} for the next\hl{ race},\hl{ regardless} of external conditions or\hl{ the}\hl{ obstacles} within themselves.\\ 
    \bottomrule
    
    \end{longtable}
\end{center}

\end{document}